\documentclass{article} 
\usepackage{iclr2021_conference,times}

\usepackage{amsmath,amsfonts,bm}

\def\eqref#1{equation~\ref{#1}}

\def\1{\bm{1}}

\newcommand{\test}{\mathcal{D_{\mathrm{test}}}}

\DeclareMathAlphabet{\mathsfit}{\encodingdefault}{\sfdefault}{m}{sl}
\SetMathAlphabet{\mathsfit}{bold}{\encodingdefault}{\sfdefault}{bx}{n}

\usepackage[linesnumbered,ruled,vlined]{algorithm2e}
\usepackage{graphicx}
\usepackage{subfig}
\usepackage{hyperref}
\usepackage{url}
\usepackage{multirow}
\usepackage{hhline}
\usepackage{booktabs}
\usepackage{cleveref}
\usepackage{enumitem}

\newenvironment{faq}{\begin{description}[style=nextline]}{\end{description}}

\title{Analyzing the Nuances of Transformers' Polynomial Simplification Abilities
}

\author{Vishesh Agarwal \\
Microsoft Vancouver\\
725 Granville St, Vancouver, BC, Canada \\
\texttt{\{visagarwal\}@microsoft.com} \\
\And
Somak Aditya \& Navin Goyal \\
Microsoft Research India \\
9 Vigyan, Lavelle Road, KA, India \\
\texttt{\{t-soadit,navingo\}@microsoft.com} \\
}

\crefformat{section}{\S#2#1#3} 
\crefformat{subsection}{\S#2#1#3}
\crefformat{subsubsection}{\S#2#1#3}

\iclrfinalcopy 
\begin{document}

\maketitle

\begin{abstract}
Symbolic Mathematical tasks such as integration often require multiple well-defined steps and understanding of sub-tasks to reach a solution. To understand Transformers' abilities in such tasks in a fine-grained manner, we deviate from traditional end-to-end settings, and explore a step-wise polynomial simplification task. Polynomials can be written in a simple normal form as a sum of monomials which are ordered in a lexicographic order. For a polynomial which is not necessarily in this normal form, a sequence of simplification steps is applied to reach the fully simplified (i.e., in the normal form) polynomial. We propose a synthetic Polynomial dataset generation algorithm that generates polynomials with unique \textit{proof} steps. Through varying coefficient configurations, input representation, proof granularity, and extensive hyper-parameter tuning, we observe that Transformers consistently struggle with numeric multiplication. We explore two ways to mitigate this: Curriculum Learning and a Symbolic Calculator approach (where the numeric operations are offloaded to a calculator). Both approaches provide significant gains over the vanilla Transformers-based baseline.

\end{abstract}

\section{Introduction}
Recently \cite{lample2019deep} showed that for the tasks of symbolic integration and solving differential equations a large number of synthetic end-to-end examples can be generated using symbolic systems. In these tasks, the authors show that Transformer networks can be trained to produce the final solution from an input integral (or differential equation) in a single step. This points to the exciting possibility of using deep neural nets to learn end-to-end theorem provers, and can be beneficial for formal mathematics \citep{Szegedy}. However, the setup combines multiple reasoning steps in a single shot: e.g., symbolic integration is a composite task and it can be non-trivial to produce intermediate steps even if the final answer is known.   
As the system in \cite{lample2019deep} is simply trained to output the top solution(s),
it is unclear what internal mechanisms enable these models to solve these problems. 
An earlier work by \cite{piotrowski2019neural} showed similar results for certain symbolic manipulation tasks and their work shares the same limitation. 

In this paper we ask if instead of only producing the end-result of symbolic manipulation or integration, can the model produce the full sequence of intermediate steps as a human-readable \emph{proof} as well. 
Inspired by \cite{piotrowski2019neural}, we explore a novel but simpler setting of \emph{polynomial simplification}. 
In this task, we begin with a polynomial which is a sum of products of factors, where each factor is again a sum of monomials (including constants). For example,
\begin{align}
\label{eq:p0}
    P_0 &=(2*x_2^2)*\overbrace{(\underbrace{3*x_2^1}_{\text{term}} + 4)}^{\text{factor}}+\overbrace{(5*x_1^2 + x_1^1*x_2^1)*(3*x_1^1)*(2)}^{\text{product}}. && \text{/* Initial */}
\end{align}
By the simplified polynomial we mean a polynomial that is written as a sum of monomials arranged in the lexicographic order. We transform a polynomial given in the above form into the simplified form in unique steps as illustrated in the example below: First, each term in a factor is simplified. Once all factors are simplified (\texttt{facstep}); then within a product, all factors are multiplied (\texttt{mulstep}). Lastly, simplified products are summed (\texttt{sumstep}). 
\begin{align}
    P_0 &=\underline{(2*x_2^2)*(3*x_2^1 + 4)}+(5*x_1^2 + x_1^1*x_2^1)*(3*x_1^1)*(2), && \textsc{/* facstep */}\\
     &= (2*x_2^2)*(3*x_2 + 4)+\underline{(5*x_1^2 + x_1^1*x_2^1)*(3*x_1^1)*(2)},&& (P_1), \textsc{/* facstep */}\\
     &= \underline{(2*x_2^2)*(3*x_2 + 4)}+(5*x_1^2 + x_1*x_2)*(3*x_1)*(2),&& (P_2), \textsc{/* mulstep */} \\
     &= (6*x_2^3 + 8*x_2^2)+\underline{(5*x_1^2 + x_1*x_2)*(3*x_1)*(2)},&& (P_3), \textsc{/* mulstep  */} \\
     &= \underline{(6*x_2^3 + 8*x_2^2)+(30*x_1^3 + 6*x_1^2*x_2)},&& (P_4), \textsc{/* sumstep */}\\
     &= 30*x_1^3 + 6*x_2^3 + 6*x_1^2*x_2 + 8*x_2^2. && (P_5), \textsc{/* Endpoint */}.
\label{eq:p1}    
\end{align}
In addition to the above setting (termed \textsc{Coarse}), we define a \textsc{Finer} setting, where a \texttt{facstep} involves simplification of a single term, a \texttt{mulstep} involves multiplications of only two factors at once, and each \texttt{sumstep} involves addition of only two products at once. 
 \cite{piotrowski2019neural} explores the task of learning symbolic re-write of an entire expression. In contrast, in our setting, for step-wise prediction, at each step the system needs to find the candidate sub-expression and a relevant simplification type to perform the simplification.  This setup resembles the traditional ATP setup where a system needs to learn and execute symbolic steps to reach a final solution. But it is much simpler, as for each step only one unique simplification is applicable. By the \textit{proof} for the simplification of the initial polynomial ($P_0$) we mean the sequence of simplification steps (leading to $P_1$ to $P_5$ in the example above). A model trained on step-wise prediction task, can be used to generate a full proof: we start with an initial polynomial, and recursively feed the model output to itself, till it generates the final simplified polynomial (in the normal form). We say that a proof is correct when all steps are correct. 
 \vspace{-0.12in}
\paragraph{Summary of our contributions.} (1) Definition of a step-wise polynomial simplification task and the dataset generation algorithm, (2) identification of bottlenecks in Transformers' math abilities validated with extensive hyperparameter search, (3) a neuro-symbolic experiment (to further understand one of these bottlenecks) where the arithmetic calculations are outsourced to a calculator, which results in significant gains across most configurations,
and (4) curriculum learning for 1-variable setting, where we observe significant gains (Table \ref{tab:cl:var1coarse} \cref{sec:appendx:curr}).

\section{Polynomial Simplification Dataset \& Task Setup}
We use the the \texttt{Sympy} library to generate the symbolic polynomials and simplification steps. To have a fine-grained control over the generated polynomials and the proof steps, we consider starting polynomials which are sums of products (e.g. $P_0$ in Eqn.~\ref{eq:p0}).\footnote{Restriction over the form of the polynomial helps us generate unique proofs, which are easier to evaluate. Detailed sampling algorithm in \cref{sec:app:alg}.}  
We randomly sample starting point polynomials. For both \textsc{coarse} and \textsc{finer} configurations, we build the proof steps as follows: (1) first we do a sequence of \texttt{facstep}s where terms get collected within a factor (such as $2x+3x$ to $5x$, $x^1$ and $1x$ become $x$); (2) then a sequence of \texttt{mulstep}s are performed where simplified factors are multiplied out; and (3) lastly, in \texttt{sumstep} simplified products are added together. The sequence of simplification steps till the endpoint constitute a full proof. The simplified endpoint polynomials are in normal form, similar to $P_5$ in Eqn.~\ref{eq:p1}.

\vspace{-0.12in}
\paragraph{Dataset Configurations.}
We vary dataset configurations along the following dimensions: (1) The number of variables in the polynomial is either 1 or 2; (2)  coefficients size: the maximum coefficients in the polynomials, products and factors resp., are varied from $\{60, 20, 5\}$ (\textsc{Small}), to $\{120, 40, 8\}$ (\textsc{Medium}) to $\{300, 100, 10\}$ (\textsc{Large}) (\textsc{Default} is $\{120, 40, 8\}$); (3) the maximum degree in polynomial and a factor has two configurations: $\{6,3\}$ (\textsc{Default}), and $\{12,5\}$ (\textsc{Medium Degree}); (4) the maximum number of terms in a simplified product and a factor 
has two configurations: $\{8,3\}$ (\textsc{Default}), and $\{20,4\}$ (\textsc{Medium Terms}). For the latter, we also set maximum products in a sum and maximum factors in a product as 5 and 4 respectively. For variation along a dimension, defaults for other dimension is used, e.g., default degree and terms for \textsc{Small Coeff}. Lastly, we try a very large configuration (\textsc{No Backtrack}) where maximum coefficients in polynomial, product and factor are $\{10125, 3375, 5\}$, maximum degree in polynomial and factor are $\{9, 3\}$; and maximum terms in a product is $27$. This is a configuration, where no sampled factor, or product is ever rejected for violating any higher-level constraint; thus capturing the effect of all constraints at once. 
As input representations, we vary between the \textit{prefix} and \textit{infix} traversals of the abstract syntax tree of the polynomial input as sequences.

\vspace{-0.12in}
\paragraph{Tasks and Metrics.} 
We identify two tasks : (1) \emph{step-wise prediction}: where an input polynomial is provided and the task is to perform the next proof step, and (2) \emph{endpoint prediction}: where given a polynomial, the task is to predict the fully simplified polynomial in one shot. 
For a fair comparison with the \emph{endpoint prediction} task, we evaluate the stepwise models on their full proof accuracy. We report the percentage of correct proofs (\textit{full proof acc}) and percentage of correct steps (\textit{stepwise acc}) using the model's top prediction (beam size 1) for every proof step .
For the step-wise prediction task the full proof accuracy is the percentage of proofs where all individual proof steps are accurate.\footnote{This is done using teacher forcing. We have attempted recursive proof generation as well, where the output from the decoder is fed to the encoder in the next step. With recursive generation, if in any step the model went wrong, it did not recover after that. Hence, \textit{proof accuracy} with teacher-forcing is a lower bound on that.} 
We also report the following: (1) error percentages grouped by each different types of steps (\texttt{facstep}, \texttt{mulstep}, and \texttt{sumstep}), (2) calibration scores of the systems based on a threshold. 
To compute the accuracy for an example (in both the tasks), we  use the \texttt{simplify} method of \texttt{Sympy} library and check symbolically whether the difference between the predicted expression and the ground-truth expression is equal to zero.  
We also calibrate the model over how confident it is of its predictions, by taking the natural log ratio of probabilities of top 2 outputs with beam 5. Whenever the ratio is greater than a threshold (usually 5), we mark that output as being \textit{sure}. Correspondingly, we report the sure rate (percentage of outputs marked \textit{sure}), precision, recall and F-1 score for calibration.

\subsection{Experimental Results} 
\label{sec:exporg}

\paragraph{Our Model.} We train a seq2seq network to predict the next proof step provided a polynomial as a sequence.  We train a Transformer network \citep{vaswani2017attention} architecture using Adam optimizer \citep{kingma2014adam}. To compare across configurations, we use a default hyperparameter setting (denoted by \textsc{Transformers-S}): 4 attention heads, 4 enc/dec layers, embedding size 256, learning rate $10^{-4}$, batch size $32$. For larger coefficient configurations, we do an exhaustive hyperparameter search varying Transformers size (S and L)\footnote{S = 4 enc/dec layers, 4 H; L = 6 enc/dec layers, 8H. Learning rates were at first: 1e-2, 5e-3, 5e-4. But,\\ lr $>5*10^{-4}$ resulted in near zero validation scores. lr $10^{-5}$ convergence took a long time.}, embedding size (256, 512), learning rate (1e-4, 5e-4, 1e-5), dropout (0, 0.5), batch-size (32, 64, 128), input representation (prefix/infix) and proof granularity (coarse/fine).
During training, we synthetically generate each batch of equations. To avoid collisions between train and test sets, we first use a fixed seed to generate the test and the validation sets of polynomial simplification full proofs and collect the simplified end-points. While sampling training batches, we make sure that the simplified versions of the input polynomial in the training batches, do not collide with any endpoints in the the test and validation set.\footnote{Authors in \cite{piotrowski2019neural} show that the probability of such collisions in the generated integration dataset by \cite{lample2019deep} to be quite high, and urge to report the test accuracy by accounting for such collisions explicitly.}  
During inference, we use beam-search with different beam widths (beam 1 and 5) to decode the expressions. For our results, beam width 1 is used for proof accuracy. Calibration results are produced using beam 5 decoding.  
During decoding, if any malformed (prefix or infix) expressions are generated, we report the percentage of such expressions.\footnote{Similar to \cite{lample2019deep}, we find that the percentage of malformed outputs was very low ($<$ 0.5\%). So we did not explicitly correct for it.} 

For the Transformers-S setting, we report the results for one and two variables for all configurations in Tables \ref{tab:var1coarse} and \ref{tab:var2coarse} (\cref{sec:trsaddl}). In Table \ref{tab:var1coarse}, we observe that \textsc{Coarse} proof-steps with \textit{Prefix} representation provides the best full proof accuracy for four out of six configurations (especially for larger coefficient sizes).  From the calibration results, we see that the winning combinations often provide the highest calibration F-1 score (more prominent for 2 variables), indicating lesser ambiguity in the decision made. As coefficient sizes grow from \textsc{Small Coeff} to \textsc{No Backtrack},  for 1 variable, the endpoint accuracy is only slightly higher ($1$ to $2\%$) than the full proof accuracy. For \textsc{Medium Terms} and \textsc{Medium Degree}, the Endpoint accuracy shows a $3.6\%$ and $13\%$ improvement respectively. For 2 variables, endpoint task accuracy is much larger in most cases. From step-wise error analysis (Appendix Tables~\ref{tab:var1error} \& \ref{tab:var2error}), shows that for 1 variable, more than 80\% of the model errors occur in the multiplication step.  In most cases, both factor simplification and addition cause close to 5\% of the model errors each. 

\begin{table}[!htb]
    \centering
\resizebox{\textwidth}{!}{%
\begin{tabular}{|l|l|c|c|c|c|c|c|c|c|c|c|c|} 
\hline
\multicolumn{1}{|c|}{\multirow{2}{*}{\begin{tabular}[c]{@{}c@{}}Polynomial\\Config\end{tabular}}} & \multicolumn{1}{c|}{\multirow{2}{*}{\begin{tabular}[c]{@{}c@{}}Proof/Input\\Format\end{tabular}}} & \multicolumn{2}{c|}{Endpoint}                                 & \multirow{2}{*}{\#Train} & \multicolumn{2}{c|}{Full Proof (Beam-1)}                                                                                    & \multicolumn{2}{c|}{Step-wise (Beam-5)}                                                                    & \multicolumn{4}{c|}{Calibration (Beam-5)}                                                     \\ 
\cline{3-4}\cline{6-13}
\multicolumn{1}{|c|}{}                                                                            & \multicolumn{1}{c|}{}                                                                             & \#EE & \begin{tabular}[c]{@{}c@{}}Endpoint\\ Acc\end{tabular} &                          & \begin{tabular}[c]{@{}c@{}}Full Proof\\ Accuracy\end{tabular} & \begin{tabular}[c]{@{}c@{}}Stepwise\\ Accuracy\end{tabular} & \begin{tabular}[c]{@{}c@{}}Top-1\\ Acc\end{tabular} & \begin{tabular}[c]{@{}c@{}}Beam-5\\ Acc\end{tabular} & \begin{tabular}[c]{@{}c@{}}Sure\\ Rate\end{tabular} & P     & R     & F1             \\ \hline \hline
Small Coeff  & Fine/Infix    & 5M                        & 96                                                                            & 4.8M                         & \textbf{98.9}                                                                      & \textbf{99.79}                                                                   & 94.46                                                                    & 95                                                                        & 92.38                                                                    & 100                    & 97.8                   & \textbf{0.99}                    \\   \hline\hline
Med Coeff &  Coarse/Prefix & 6.1M                      & 95.87                                                                         & 5.3M                         & \textbf{93.6}                                                                      & \textbf{98.58}                                                                   & 86.6                                                                     & 88.47                                                                     & 82.83                                                                    & 99.88                  & 95.54                  & \textbf{0.98}                                        \\ \hline\hline
Large Coeff  & Coarse/Prefix & 6.5M                      & 85.87                                                                         & 3.5M                         & \textbf{83.5}                                                                      & \textbf{96.25}                                                                   & 80.6                                                                     & 83.3                                                                      & 75                                                                       & 99.91                  & 92.97                  & 0.96                    \\                      \hline\hline
No BT  & Coarse/Prefix & 6.6M                      & 78.87                                                                         & 5.6M                         & \textbf{79.7}                                                                      & \textbf{95.38}                                                                   & 81.93                                                                    & 85.57                                                                     & 72.2                                                                     & 100                    & 88.12                  & 0.94                    \\                   \hline\hline
Med Deg & Coarse/Infix  & 9.2M                      & 96.4                                                                          & 4.9M                         & \textbf{92.8}                                                                      & \textbf{98.26}                                                                   & 87.18                                                                    & 88.96                                                                     & 81.12                                                                    & 100                    & 93.05                  & 0.96                                       \\ \hline\hline
Med Terms   & Coarse/Prefix & 7M                        & 89.8                                                                          & 4.3M                         & \textbf{76.3}                                                                      & \textbf{95.78}                                                                   & 87.8                                                                     & 91.17                                                                     & 81.3                                                                     & 100                    & 92.6                   & \textbf{0.96}                    \\ \hline
\end{tabular}%
}
    \caption{\textsc{Transformers-S} results for 1 variable. Only highest proof accuracy combinations are shown for each coefficient configuration. Full results in Appendix.}
    \label{tab:var1coarse}
\end{table}

\begin{table}[!htb]
    \centering
\resizebox{\textwidth}{!}{%
\begin{tabular}{|l|l|c|c|c|c|c|c|c|c|c|} 
\hline
\multicolumn{1}{|c|}{\multirow{2}{*}{\begin{tabular}[c]{@{}c@{}}Polynomial\\Config\end{tabular}}} & \multicolumn{1}{c|}{\multirow{2}{*}{\begin{tabular}[c]{@{}c@{}}Proof/Input\\Format\end{tabular}}} & \multirow{2}{*}{\#Train} & \multicolumn{2}{c|}{Full Proof (Beam-1)}                                                                                     & \multicolumn{2}{c|}{Step-wise (Beam-5)}                                                                    & \multicolumn{4}{c|}{Calibration (Beam-5)}                                                                                                                                                    \\ 
\cline{4-11}
\multicolumn{1}{|c|}{}                                                                            & \multicolumn{1}{c|}{}                                                                             &                          & \begin{tabular}[c]{@{}c@{}}Full Proof \\ Accuracy\end{tabular} & \begin{tabular}[c]{@{}c@{}}Stepwise\\ Accuracy\end{tabular} & \begin{tabular}[c]{@{}c@{}}Top-1 \\ Acc\end{tabular} & \begin{tabular}[c]{@{}c@{}}Beam-5 \\ Acc\end{tabular} & \begin{tabular}[c]{@{}c@{}}Sure\\ Rate\end{tabular} & P     & R     & F1    \\ 
\hline \hline
  \multirow{2}{*}{\begin{tabular}[c]{@{}l@{}}Large\\ Coeff\end{tabular}}                                                                        & h/infix    &   3.6M     & 84.6                                                      & 96.79                                                  & 96.5                                                 & 98.02                                                 & 91.62                                               & 99.93 & 94.88 & 0.97 \\ \cline{2-11}\cline{2-11} 
                                                                          & \texttt{cal}/infix  &  3.2M      & \textbf{95.7}                                                      & \textbf{99.16}                                                  & 97.98                                                & 98.72                                                 & 95.58                                               & 99.94 & 97.49 & 0.99 \\ \hline\hline
\multirow{2}{*}{\begin{tabular}[c]{@{}l@{}}No\\ Backtrack\end{tabular}}                                                                          & h/infix    &   3.2M     & 82.6                                                      & 96.85                                                  & 95.98                                                & 97.86                                                 & 88.42                                               & 99.98 & 92.1  & 0.96 \\ \cline{2-11}\cline{2-11} 
 \cline{2-11} 
                                                                          & \texttt{cal}/infix  & 3.8M   & \textbf{87.2}                                                      & \textbf{97.94}                                                  & 97.26                                                & 98.56                                                 & 94.18                                               & 99.94 & 96.77 & 0.98 \\ \hline\hline
 \multirow{2}{*}{\begin{tabular}[c]{@{}l@{}}Medium\\ Degree\end{tabular}}        & h/infix    &   3.3M     & \textbf{90.8}      & 97.81   & 96.66          & 98.14   & 90.34  & 99.93 & 93.4  & 0.97 \\ \cline{2-11} \cline{2-11} 
                                                                          & \texttt{cal}/infix  &   2.8M     & \textbf{90.5}                                                      & \textbf{98.08 }                                                 & 97.92                                                & 98.94                                                 & 92.86                                               & 99.96 & 94.79 & 0.97 \\ \hline\hline
\multirow{2}{*}{\begin{tabular}[c]{@{}l@{}}Medium\\ Terms\end{tabular}}                                                                          & h/infix    &   3.5M     & 79.5                                                      & 96.49                                                  & 96.16                                                & 97.92                                                 & 91.19                                               & 100   & 94.82 & 0.97 \\ \cline{2-11} \cline{2-11} 
 \cline{2-11} & \texttt{cal}/infix  & 3.7M & \textbf{89.5} & \textbf{98.45} & 97.88 &  99.08 & 94.88 & 99.92 & 96.85 & 0.98 \\ \hline
\end{tabular}%
}
    \caption{1 variable, hyperparmeter-tuning and symbolic calculator experiments.
    \textit{h/*}  rows denote the performance of models with best validation scores over all explored hyperparameter configurations (including input representation variations). \texttt{cal}/* provides the same for the calculator setting for the best models after hyperparameter search. Full table in Appendix.
    }
    \label{tab:var1combo}
\end{table}


\paragraph{Symbolic Calculator.}
As 80\% of the errors occurred in multiplication step, we separately tested the Transformer's ability to do arithmetic, by creating datasets involving multiplication and addition of 4-digit and 9-digit numbers. While the models quickly achieved an accuracy of 99\% for addition; for multiplication, they could not go beyond even 1\% after seeing 2M examples. Hence, we devise a setting where polynomial simplification steps only involve symbolic addition and multiplication, without any arithmetic manipulation. For example, for given expressions $(3*x_1)*(4*x_1)$, the model is trained to output $[3*4]*x_1^2$, where ``$[\cdot]$'' signifies that a calculator needs to simplify the inner expression. In Tab.~\ref{tab:var1combo}, we report results for both the standard and the calculator setting. Each row denotes the test-set results for the model with best validation score\footnote{Full table and best hyperparameter settings per row in Appendix Table \ref{tab:var1comboerrors}.}, for a fixed dataset configuration and input representation. As shown, in many configurations there is a significant increase in proof accuracy (11\% in Large Coeff and Medium Terms) for both prefix and infix representation.  

\paragraph{Curriculum Learning.} For this task, the sub-tasks (addition, multiplication) and their dependencies are well-defined. To exploit such dependencies, we explore different curriculum strategies based on the Mastering-Rate-based (MR) curriculum learning algorithm by \cite{willems2020mastering}. In this work, authors define the task dependencies as a graph, and a sub-task is only sampled from (or learnt) when its parent tasks are \emph{mastered}. From our experiments (detailed in Appendix), we observe that in 1-variable setting, as coefficient size grows from \textsc{Small}, \textsc{Medium}, \textsc{Large} to \textsc{No Backtrack} - the improvements in full proof accuracy steadily increase from $1\%$ to $10.84\%$. 


\paragraph{Conclusion and Results Summary.}  We define the polynomial simplification task and present the synthetic dataset generation procedure. A default Transformers settings show that for smaller configurations, Transformers learn to generate whole proofs with very high accuracy.
For large configurations, even with exhaustive hyper-parameter tuning, even a larger Transformer model suffers. 
We identify that arithmetic multiplication is a consistent bottleneck for Transformers. Through a neuro-symbolic model, where numeric operations are outsourced to a calculator, we observe high gains on full proof accuracy. Lastly, we observe carefully designed curricula can also boost full proof accuracy up to 10\% for large coefficient sizes. These results indicate some potential ingredients that could be useful for designing very high accuracy models.


\bibliography{polysimp}
\bibliographystyle{iclr2021_conference}

\appendix

\section{Polynomial Simplification Dataset Creation and The Sampling Algorithm}
\label{sec:app:alg}
We proceed similarly to \cite{lample2019deep} to generate the symbolic polynomials and simplified steps synthetically using the \texttt{Sympy} library of Python. To have a fine-grained control over the generated polynomials and well-defined proof steps, we consider polynomials which are sums of products\footnote{The generation algorithm in \cite{lample2019deep} may generate nested sums and products. For such polynomials, an unique proof sequence is hard to define which makes whole \textit{proof}s harder to evaluate. Our restriction over the form of the polynomial helps us generate unique proofs, which are easier to evaluate.}. 
We also note that symbolic generation using the \texttt{Sympy} library lets us ensure correctness of each generated expressions and validity of each steps. 

\subsection{Notations}
We start with the set of variables $x_\mathcal{P} =\{x_1,\ldots,x_{\mathrm{nvar}} \}$.  We  represent the starting point polynomial $\mathcal{P}_0$ in $x_\mathcal{P}$ as the sum of products of factors: 
\begin{equation}
  \begin{aligned} 
\mathcal{P}_0 &=  P_1 + P_2 + \ldots +P_{\mathrm{nprod}}, \\ 
P_i &=  \prod_{j=1}^{\mathrm{nfac}_i} f_{ij}, 
\label{eq1}
\end{aligned}  
\end{equation}
where  each factor ($f_{ij}$)  has the form $f = \sum_{k}^{} ({a_k * \prod_{l}^{} x_{kl}^{d_{kl}}})$, where $~x_{kl} \in x_\mathcal{P}$ (dropping $i,j$ for clarity). Here coefficients $ a_{k} \in \mathbb{N}^+$,
and powers of the variables $d_{kl} \in \mathbb{N}$. $\mathrm{nprod}$ is the number of products and $\mathrm{nfac}_i$ denotes the number of factors in $P_i$.\\ We denote the set of factors as $f_\mathcal{P} =\{f_{ij} | \exists i, P_i =  \prod_{j=1}^{\mathrm{nfac}_i} f_{ij}\}$.
The simplified endpoint polynomial is of the form $\mathcal{\hat{P}} = \sum_{m=1}^{q} \hat{t}_m$,  where $\hat{t}_m = \hat{a}_m * \prod_{n}^{} {x_n}^{d_{mn}}$, \text{ where } ~$x_n \in x_\mathcal{P}$. We use the symbol $\hat{P}_i$ to denote the simplified form of $P_i$. The functions $\mathrm{terms}(), \mathrm{vars}(), \mathrm{coeffs}()$ returns a list of terms, variables, coefficients in the input expression. Our sampling algorithm guarantees that the generated polynomial and its simplified endpoint abides by constraints on number of terms, products, factors and variables; limit on degree and coefficient sizes. An example is $\mathrm{nprod} \in  \{2, \ldots, \mathrm{maxP_\mathcal{P}}\}$. We provide the full list of constraints and notations in Table \ref{tab:constraints}.

\begin{table}[!ht]
\centering
\begin{tabular}{@{}lll@{}}
\toprule
\begin{tabular}[c]{@{}l@{}}Term \\ Constraints\end{tabular}        & \begin{tabular}[c]{@{}l@{}}\#Products\\ \#Factors in $P_i$\\ \#Terms in $f_{ij}$\\ \#Terms in $\hat{P}_i$\end{tabular} & \begin{tabular}[c]{@{}l@{}}$\mathrm{nprod} \in  \{2, \ldots, \mathrm{maxP_\mathcal{P}}\}$\\ $\mathrm{nfac}_i \in \{2, \ldots, \mathrm{maxf_P}\}, \forall i \in \{1,\ldots,\mathrm{nprod}\}$\\ $|\mathrm{terms}(f_i)|   \in \{1, \ldots, \mathrm{maxT_f}\}, \forall f_ij \in f_\mathcal{P}$\\ $|\mathrm{terms}(\hat{P_i})| \leq \mathrm{maxT_P} \forall P_i \in \mathcal{P}_0$\end{tabular} \\ \midrule
\begin{tabular}[c]{@{}l@{}}Degree \\ Constraints\end{tabular}      & \begin{tabular}[c]{@{}l@{}}\#Degree in $\hat{\mathcal{P}}$\\ \#Degree in $f_{ij}$\end{tabular}                            & \begin{tabular}[c]{@{}l@{}}$\sum_{}^{}d_{mn} \leq \mathrm{D_\mathcal{P}},~\forall m~ \hat{t_m} \in \mathrm{terms}(\hat{P}), \forall n~ x_n \in \mathrm{vars}(\hat{t_m})$\\ $\sum_{}^{}d_{kl} \leq \mathrm{D_f}, \forall k~ \mathrm{terms}(f_{ij}), \forall f_{ij} \in f_\mathcal{P}$\end{tabular}                                                                                                            \\ \midrule
\begin{tabular}[c]{@{}l@{}}Variable \\ Constraints\end{tabular}    & \begin{tabular}[c]{@{}l@{}}\#Variables in $\mathcal{P}_0$\\ \#Variables in $P_i$\\
\#Variables in $f_i$\end{tabular}                            & \begin{tabular}[c]{@{}l@{}}$|x_\mathcal{P}| \leq \mathrm{V_\mathcal{P}}$\\ $|\mathrm{vars(P_i)}| \leq \mathrm{V_P}, \forall P_i \in \mathcal{P}_0$ \\
$|\mathrm{vars(f_{ij})}| \leq \mathrm{V_f}, \forall f_j \in f_\mathcal{P}$\end{tabular}                                                                                                                                                                                                                          \\ \midrule
\begin{tabular}[c]{@{}l@{}}Coefficient \\ Constraints\end{tabular} & \begin{tabular}[c]{@{}l@{}}Coeff in $\hat{\mathcal{P}}$\\ Coeff in $\hat{P}_i$\\ Coeff in $f_i$\end{tabular}           & \begin{tabular}[c]{@{}l@{}}$\hat{a_j} \leq \mathrm{C_\mathcal{P}, } \forall \hat{a_j} \in \mathrm{coeffs}(\hat{P})$\\ $\hat{a_{ij}} \leq \mathrm{C_P}, \forall a~ \mathrm{coeffs}(\hat{P_i}), \forall P_i \in \mathcal{P}_0$\\ $a_k \leq \mathrm{C_f}, \forall a~ \mathrm{coeffs}(f_{ij}), \forall f_{ij} \in f_\mathcal{P}$\end{tabular}                                                    \\ \bottomrule
\end{tabular}
\caption{List of notations, and corresponding constraints that a generated polynomial satisfies.}
\label{tab:constraints}
\end{table}

\subsection{Building a Polynomial Proof}
 Here, we briefly describe the starting polynomial generation process; detailed algorithm is in the appendix. Any randomly sampled polynomial (represented as a sum of products) can be included as a starting point in the dataset as long as the polynomial respects certain configuration parameters (in Appendix Table \ref{tab:constraints}). This is unlike \cite{lample2019deep}, where many randomly generated integrals (or differential equations) might not have a solution. 
Hence, we randomly sample the constraint parameters in a top-down manner; and then construct terms, factors and products in a bottom-up manner using the parameters. We first sample the following 1) a set of participating variables ( $x_\mathcal{P}$), 2) maximum degree for any monomial in the simplified polynomial ($\mathrm{mdeg}$), and 3) the number of products in the starting polynomial ($\mathrm{nprod}$). We then call the algorithm \texttt{buildProduct} (Algorithm \ref{alg1:prod} in appendix) to create $\mathrm{nprod}$ individual products. 
    
\begin{algorithm}[!htpb]
\DontPrintSemicolon
\SetAlgoLined
\SetNoFillComment
\SetKwInOut{Limit}{Constraints}
\KwIn{$x_{\mathcal{P}}$, \texttt{mdeg}}
\Limit{\texttt{nvars\_prod, max\_coeff\_prod, max\_fac\_prod, max\_terms\_prod}}
\KwOut{A list of factors $F_{seq}$}
 Sample $nvar$ $\in \{\texttt{num\_vars\_fac}, \ldots, \texttt{nvars\_prod}\}$ \;
 $nvar = min(|x_{\mathcal{P}}|, nvar)$\;
 Sample $nvar$ variables from $x_{\mathcal{P}}$ as $x_{\mathcal{P}_i}$\ \tcp*{Variable set for this product}
 Sample $nfac$ $ \in \{2, \ldots, \texttt{max\_fac\_prod}\}$\ \tcp*{\#Factors for this product}
 \tcc{Get maximum degree, terms and coefficient available}\
 $rdegree$ = $mdeg$, $rterms$ = \texttt{max\_terms\_prod}, $rcoeff$ = \texttt{max\_coeff\_prod}\;
 $cprod$ = 1 \tcp*{Keeping track of product built till now}
 $F_{seq} = [~]$\;
 \For{$j\gets1$ \KwTo $nfac$}{
  $f_j = \texttt{buildFactor}(x_{\mathcal{P}_i}, rdegree, rterms, rcoeff)$\;
    \tcc{Update degree, terms and coefficient for next factor}\
  $cprod = cprod * f_j$\;
  $rdegree = rdegree - degree(f_j)$\; 
  $rterms$ = \texttt{max\_terms\_prod}/$|terms(cprod)|$\; 
  $rcoeff$ = \texttt{max\_coeff\_prod}/$max(coeffs(cprod))$\; 
  Append $f_j$ in $F_{seq}$\;
  \If{$rdegree == 0$}{
      break\;
  }
  }
  Shuffle $F_{seq}$
 \caption{BuildProduct (Sampling Products)}
 \label{alg1:prod}
\end{algorithm}

\paragraph{Building a Product} In \texttt{buildProduct} (Algorithm \ref{alg1:prod}), first we sample $\mathrm{nfac}_i$, the maximum number of factors in the product ($P_i$). We then build factors sequentially.  For each new factor, we sample a subset of variables in a factor.
We pass on product-level constraints such as maximum degree in a product, maximum terms in a product, and maximum coefficient for a product as $\mathrm{rdegree}, \mathrm{rterms}$ and $\mathrm{rcoeff}$ respectively; and call the sub-routine \texttt{buildFactor} (Algorithm \ref{alg2:fac} to create a factor. After a factor is sampled, the constraints $\mathrm{rdegree}, \mathrm{rterms}$ and $\mathrm{rcoeff}$ are updated. \texttt{buildFactor} is used to create at most $\mathrm{nfac}_i$ factors, that all abide by the above constraints and stops if the limit of maximum degree in the product is reached. The terms in a factor are arranged in a lexicographical order. Since, this sequential generation of factors may induce a certain pattern of decreasing degrees and coefficients, we shuffle the factors to create the final product.

\paragraph{Simplification Steps and Full Proof}
For both \textsc{coarse} and \textsc{finer} configurations, we build the \textit{proof} steps in the following way: 1) first we do a sequence of \texttt{facstep}s where terms get collected within a factor (such as $2x+3x$ to $5x$, $x^1$ and $1x$ becomes $x$); 2) then a sequence of \texttt{mulstep}s are performed where simplified factors are multiplied out; and 3) lastly, in \texttt{sumstep} simplified products are added together. As mentioned before, the sequence of simplification steps till the endpoint constitute a full \textit{proof}.

The polynomial sampling algorithms  \texttt{buildProduct} and \texttt{buildFactor} are provided in  Algorithms \ref{alg1:prod} and \ref{alg2:fac} respectively.

\begin{algorithm}[!htpb]
{
\DontPrintSemicolon
\SetAlgoLined
\SetNoFillComment
\SetKwInOut{Limit}{Constraints}
\KwIn{$x_{\mathcal{P}_i}, rdegree, rterms, rcoeff$}
\Limit{\texttt{num\_vars\_fac, max\_coeff\_fac, max\_terms\_fac, max\_degree\_fac}}
\KwOut{A factor $f_j$, Number of terms $nterms_j$}
 Sample $nvar$ $\in \{1, \ldots, \texttt{num\_vars\_fac}\}$ \;
 $cvars$ = Sample $nvar$ variables from $x_{\mathcal{P}_i}$ \tcp*{Variable set for this factor}
 Sample $nterms \in \{1, \ldots, min(\texttt{max\_terms\_fac}, rterms)\}$\; \tcp*{\# Terms for this factor}
 Sample $\{d_k\}^{nterms}_{k=1}$, s.t. $d_k \in \{0,\ldots, min(\texttt{max\_degree\_fac}, rdegree)\}$ \; \tcp*{Term degrees: degree 0 allows for constant terms}
 Sample $\{c_k\}^{nterms}_{k=1}$, s.t. $c_k \in \{1, \ldots, min(\texttt{max\_coeff\_fac}, rcoeff)\}$\; \tcp*{Term coefficients}
 \For{$k\gets1$ \KwTo $nterms$}{
     selects $d[k]$ variables from $cvars$ with replacement\; \tcp*[l]{ E.g. if $d[k] = 4, cvars = [x_1, x_2]  \text{. May sample } [x_1,x_2, x_1, x_1]$}
     Convert the selected $d[k]$ variables to a term\tcp*[l]{ $t_k = c_k * x_1^3 * x_2$,}
 }
 $f_j = \sum_{k=1}^{nterms} t_k$\;
 }
  return $f_j$\;
\caption{BuildFactor (Sampling A Factor)}
\label{alg2:fac}
\end{algorithm}

\section{Result Table: FAQs}
\label{sec:faq}
\begin{faq}
\item[Q: How do you generate a proof?]
We feed each input step to the model, and get the next step using greedy decoding.
\item[Q: When is a proof marked correct?]
When all the intermediate steps generated and the final simplified polynomial match the ground truth. These results are recorded in the full proof major column.
\item[Q: When is the step-wise major column then?]
We also tried using beam-5 for decoding. The results for that are summarised, per step, in the step-wise major column.
\item[Q: Why are "full proof $>$ stepwise accuracy" and "step-wise $>$ top-1 accuracy" different?]
As explained earlier, full proof results are with greedy decoding, and step wise are with beam-5. Since the top output for beam-5 can be different from greedy decoding, we have these two different metrics. As one can see, greedy decoding usually does better than beam search.
\item[Q: How are "step-wise $>$ top-1 accuracy" and "step-wise $>$ beam-5 accuracy" different?]
Both of those metrics are for beam-5 decoding. Top-1 accuracy measures how many times the top outputs in the beam matches the ground truth, whereas beam-5 accuracy measures how many times \textit{ANY} output in the beam matches ground truth. 
\item[Q: What is calibration?]
Calibration is used to quantify how confident a model of its predictions. If the top prediction of a model also has a very low probability, we can say that the model is not able to come up with a good answer. We use the log ratio of model probability of top 2 outputs with beam-5 decoding to measure how sure model is of its top prediction. We call this ratio the confidence score of the model.
\item[Q: Why use the log \textit{ratio of top 2}?]
Note that in our configuration, only one output is correct for each input. In our experiments we found that when the model found the correct output, it gave a very high score to it, and low scores for rest of the beam predictions. Whereas, when it gave incorrect answers, the top few beam outputs had similar (and low) scores. So, we use the log ratio measure for confidence.
\item[Q: How do you know when the confidence score is enough to mark an output as sure?]
We used AUC-ROC and AUPRC to calibrate a threshold for the model calibration scores. An output with score above the threshold is marked "sure" and below that is marked "unsure".
\item[Q: What are the subcolumns under the calibration major column?]
Sure rate is the \%age of outputs marked as correct. Precision measures how many sure outputs are correct. Recall measures how many of the correct outputs were marked as sure. The F-1 score is calculated using that precision and recall.
\item[Q: Finally, what are \#EE and \#Train?]
To compare the sample efficiency with the baseline models, we note the number of examples the models see before convergence. Note that for the \textsc{ENDPOINT} model, an example consists of an initial polynomial and final simplified version pair. Whereas for the stepwise models, one example consists of one proof step (NOT the whole proof).
\end{faq}

\section{Problem Space Size Estimation}
\label{sec:psize}
For smaller configurations, it is probable that eventually all simplified polynomials would be included in the training data. To account for this, we estimate the problem space size for each configuration and report the size of training data for comparison. 
We randomly generate two sets of starting polynomials say $S_1$ and $S_2$, and check for collisions among them. Assuming the actual size is $X$ and uniform distribution over all starting polynomials, the expected number of collisions would be $R = \frac{S_1 *S_2}{X}$. Using the above method, we  estimate the number of un-simplified polynomials and
the number of unique endpoints, and report in Table \ref{tab:colide}. We observe that compared to the number of training examples it took for the models to converge in both End-point and Step-wise prediction tasks, the space of possible equations is often 25 (or more) times higher. 

Sampled polynomials are not uniformly distributed as we assign an equal probability while sampling polynomials of lower and higher degrees, say 3 and 6; whereas there are more polynomials of degree 6 than degree 3. For non-uniform distributions, we expect more collisions as higher probability equations are more likely to occur in both $S_1$ and $S_2$. Moreover, since many equations may map to the same endpoint, such collisions for endpoints are even more likely. 
Thus, our empirical estimate of the population size provides a lower bound on the true value.

\begin{table}[!htpb]
\centering
\resizebox{0.9\textwidth}{!}{%
\begin{tabular}{|c|c|c|c|c|}
\hline
\multirow{2}{*}{\textbf{Config}} & \multicolumn{2}{c|}{NVAR = 1}                                                                                                       & \multicolumn{2}{c|}{NVAR = 2}                                                                                                       \\ \cline{2-5} 
                                 & \begin{tabular}[c]{@{}c@{}}Equation\\ Size Estimate\end{tabular} & \begin{tabular}[c]{@{}c@{}}Endpoint\\ Size Estimate\end{tabular} & \begin{tabular}[c]{@{}c@{}}Equation\\ Size Estimate\end{tabular} & \begin{tabular}[c]{@{}c@{}}Endpoint\\ Size Estimate\end{tabular} \\ \hline
\textbf{SMALL COEFF}             & 104M                                                             & 8.24M                                                            & 184M                                                             & 27.4M                                                            \\ \hline
\textbf{MEDIUM COEFF}            & 179M                                                             & 16.3M                                                            & 325M                                                             & 42.4M                                                            \\ \hline
\textbf{LARGE COEFF}             & 289M                                                             & 32M                                                              & 507M                                                             & 68.8M                                                            \\ \hline
\textbf{NO BACKTRACK}            & 324M                                                             & 54.9M                                                            & 538M                                                             & 104M                                                             \\ \hline
\textbf{MEDIUM DEG}              & 459M                                                             & 67.4M                                                            & 902M                                                             & 144M                                                             \\ \hline
\textbf{MEDIUM TERMS}            & 866M                                                             & 31.5M                                                            & 1.73B                                                            & 801M                                                             \\ \hline
\end{tabular}%
}
\caption{Size Estimates for the problem space, after generating sets of size 5M.}
\label{tab:colide}
\end{table}

\section{Input Representation (Additional Results)} 
\label{sec:trsaddl}
For \textsc{Transformers-S} setting, we present the complete results for 1 variable in Table \ref{tab:var1coarsefull}. For 2 variables, the \textsc{Coarse} configuration results are in Table \ref{tab:var2coarse} and \textsc{FINE} configuration in Table \ref{tab:results2varfine}. The errors made by the models for 1 Variable and 2 Variable settings are presented in Tables \ref{tab:var1error} and \ref{tab:var2error} respectively.

\begin{table}[!htpb]
    \centering
\resizebox{\textwidth}{!}{%
\begin{tabular}{|l|l|c|c|c|c|c|c|c|c|c|c|c|} 
\hline
\multicolumn{1}{|c|}{\multirow{2}{*}{\begin{tabular}[c]{@{}c@{}}Polynomial\\Config\end{tabular}}} & \multicolumn{1}{c|}{\multirow{2}{*}{\begin{tabular}[c]{@{}c@{}}Proof/Input\\Format\end{tabular}}} & \multicolumn{2}{c|}{Endpoint}                                 & \multirow{2}{*}{\#Train} & \multicolumn{2}{c|}{Full Proof (Beam-1)}                                                                                    & \multicolumn{2}{c|}{Step-wise (Beam-5)}                                                                    & \multicolumn{4}{c|}{Calibration (Beam-5)}                                                     \\ 
\cline{3-4}\cline{6-13}
\multicolumn{1}{|c|}{}                                                                            & \multicolumn{1}{c|}{}                                                                             & \#EE & \begin{tabular}[c]{@{}c@{}}Endpoint\\ Acc\end{tabular} &                          & \begin{tabular}[c]{@{}c@{}}Full Proof\\ Accuracy\end{tabular} & \begin{tabular}[c]{@{}c@{}}Stepwise\\ Accuracy\end{tabular} & \begin{tabular}[c]{@{}c@{}}Top-1\\ Acc\end{tabular} & \begin{tabular}[c]{@{}c@{}}Beam-5\\ Acc\end{tabular} & \begin{tabular}[c]{@{}c@{}}Sure\\ Rate\end{tabular} & P     & R     & F1             \\ \hline \hline
\multirow{4}{*}{\begin{tabular}[c]{@{}l@{}}Small \\ Coeff\end{tabular}}  & Coarse/Infix  & 5M                        & 96                                                                            & 3.6M                         & 95                                                                                 & 98.83                                                                            & 88.13                                                                    & 89.67                                                                     & 83.2                                                                     & 100                    & 94.4                   & 0.97                    \\ \cline{2-13} 
                                                                         & Fine/Infix    & 5M                        & 96                                                                            & 4.8M                         & \textbf{98.9}                                                                      & \textbf{99.79}                                                                   & 94.46                                                                    & 95                                                                        & 92.38                                                                    & 100                    & 97.8                   & \textbf{0.99}                    \\ \cline{2-13} 
                                                                         & Coarse/Prefix & 5.2M                      & 97.8                                                                          & 3.2M                         & 95.3                                                                               & 98.97                                                                            & 87.83                                                                    & 89.37                                                                     & 83.03                                                                    & 100                    & 94.54                  & 0.97                    \\ \cline{2-13} 
                                                                         & Fine/Prefix   & 5.2M                      & 97.8                                                                          & 4.4M                         & 96.9                                                                               & 99.4                                                                             & 95.1                                                                     & 95.83                                                                     & 93.13                                                                    & 99.96                  & 97.9                   & \textbf{0.99}                    \\ \hline\hline
\multirow{4}{*}{\begin{tabular}[c]{@{}l@{}}Medium \\ Coeff\end{tabular}} & Coarse/Infix  & 4.1M                      & 91.2                                                                          & 4.3M                         & 92.8                                                                               & 98.24                                                                            & 88.97                                                                    & 91.67                                                                     & 84.3                                                                     & 100                    & 94.75                  & 0.97                    \\ \cline{2-13} 
                                                                         & Fine/Infix    & 4.1M                      & 91.2                                                                          & 2.9M                         & 90.3                                                                               & 97.99                                                                            & 86.1                                                                     & 87.68                                                                     & 81.14                                                                    & 100                    & 94.24                  & 0.97                    \\ \cline{2-13} 
                                                                         & Coarse/Prefix & 6.1M                      & 95.87                                                                         & 5.3M                         & \textbf{93.6}                                                                      & \textbf{98.58}                                                                   & 86.6                                                                     & 88.47                                                                     & 82.83                                                                    & 99.88                  & 95.54                  & \textbf{0.98}                    \\ \cline{2-13} 
                                                                         & Fine/Prefix   & 6.1M                      & 95.87                                                                         & 4.5M                         & 91.7                                                                               & 98.37                                                                            & 95.1                                                                     & 96.43                                                                     & 91.27                                                                    & 100                    & 95.97                  & \textbf{0.98}                    \\ \hline\hline
\multirow{4}{*}{\begin{tabular}[c]{@{}l@{}}Large \\ Coeff\end{tabular}}  & Coarse/Infix  & 4.8M                      & 83.73                                                                         & 3.4M                         & 82.1                                                                               & 95.97                                                                            & 92.34                                                                    & 94.22                                                                     & 87.2                                                                     & 99.98                  & 94.41                  & 0.97                    \\ \cline{2-13} 
                                                                         & Fine/Infix    & 4.8M                      & 83.73                                                                         & 3.4M                         & 82.5                                                                               & 96.44                                                                            & 92.32                                                                    & 94.26                                                                     & 87.5                                                                     & 99.98                  & 94.76                  & 0.97                    \\ \cline{2-13} 
                                                                         & Coarse/Prefix & 6.5M                      & 85.87                                                                         & 3.5M                         & \textbf{83.5}                                                                      & \textbf{96.25}                                                                   & 80.6                                                                     & 83.3                                                                      & 75                                                                       & 99.91                  & 92.97                  & 0.96                    \\ \cline{2-13} 
                                                                         & Fine/Prefix   & 6.5M                      & 85.87                                                                         & 3.2M                         & 82                                                                                 & 96.32                                                                            & 79.13                                                                    & 80.63                                                                     & 75.57                                                                    & 99.96                  & 95.45                  & \textbf{0.98}                    \\ \hline\hline
\multirow{4}{*}{\begin{tabular}[c]{@{}l@{}}No\\ Backtrack\end{tabular}}  & Coarse/Infix  & 5.9M                      & 80.1                                                                          & 3.8M                         & 75.6                                                                               & 94.62                                                                            & 72.74                                                                    & 77.28                                                                     & 61.8                                                                     & 99.9                   & 84.88                  & 0.92                    \\ \cline{2-13} 
                                                                         & Fine/Infix    & 5.9M                      & 80.1                                                                          & 4M                           & 74.5                                                                               & 94.76                                                                            & 88.34                                                                    & 90.9                                                                      & 79.44                                                                    & 99.92                  & 89.86                  & 0.95                    \\ \cline{2-13} 
                                                                         & Coarse/Prefix & 6.6M                      & 78.87                                                                         & 5.6M                         & \textbf{79.7}                                                                      & \textbf{95.38}                                                                   & 81.93                                                                    & 85.57                                                                     & 72.2                                                                     & 100                    & 88.12                  & 0.94                    \\ \cline{2-13} 
                                                                         & Fine/Prefix   & 6.6M                      & 78.87                                                                         & 4.2M                         & 74.7                                                                               & 95.23                                                                            & 79                                                                       & 82.03                                                                     & 72.23                                                                    & 100                    & 91.43                  & \textbf{0.96 }                   \\ \hline\hline
\multirow{4}{*}{\begin{tabular}[c]{@{}l@{}}Medium\\ Degree\end{tabular}} & Coarse/Infix  & 9.2M                      & 96.4                                                                          & 4.9M                         & \textbf{92.8}                                                                      & \textbf{98.26}                                                                   & 87.18                                                                    & 88.96                                                                     & 81.12                                                                    & 100                    & 93.05                  & 0.96                    \\ \cline{2-13} 
                                                                         & Fine/Infix    & 9.2M                      & 96.4                                                                          & 3.3M                         & 83.4                                                                               & 96.12                                                                            & 88.26                                                                    & 90.44                                                                     & 83.04                                                                    & 99.95                  & 94.04                  & \textbf{0.97}                    \\ \cline{2-13} 
                                                                         & Coarse/Prefix & 7M                        & 94.33                                                                         & 4.3M                         & 87.7                                                                               & 96.82                                                                            & 77.33                                                                    & 82.13                                                                     & 69.33                                                                    & 100                    & 89.66                  & 0.95                    \\ \cline{2-13} 
                                                                         & Fine/Prefix   & 7M                        & 94.33                                                                         & 5.9M                         & 90.6                                                                               & 97.92                                                                            & 82.2                                                                     & 83.7                                                                      & 77.27                                                                    & 99.96                  & 93.96                  & \textbf{0.97}                    \\ \hline\hline
\multirow{4}{*}{\begin{tabular}[c]{@{}l@{}}Medium\\ Terms\end{tabular}}  & Coarse/Infix  & 4.6M                      & 81.9                                                                          & 2.3M                         & 72.7                                                                               & 93.99                                                                            & 79.44                                                                    & 82.22                                                                     & 68.22                                                                    & 99.97                  & 85.85                  & 0.92                    \\ \cline{2-13} 
                                                                         & Fine/Infix    & 4.6M                      & 81.9                                                                          & 2.8M                         & 75.1                                                                               & 95.42                                                                            & 86.2                                                                     & 88.48                                                                     & 76.72                                                                    & 99.92                  & 88.93                  & 0.94                    \\ \cline{2-13} 
                                                                         & Coarse/Prefix & 7M                        & 89.8                                                                          & 4.3M                         & \textbf{76.3}                                                                      & \textbf{95.78}                                                                   & 87.8                                                                     & 91.17                                                                     & 81.3                                                                     & 100                    & 92.6                   & \textbf{0.96}                    \\ \cline{2-13} 
                                                                         & Fine/Prefix   & 7M                        & 89.8                                                                          & 3.2M                         & 74.8                                                                               & 95.55                                                                            & 87.67                                                                    & 90.4                                                                      & 76.9                                                                     & 100                    & 87.72                  & 0.93                    \\ \hline
\end{tabular}%
}
    \caption{Results for 1 variable in the \textsc{Coarse} and \textsc{Fine} configuration for both Infix and Prefix representation.}
    \label{tab:var1coarsefull}
\end{table}

\begin{table}[!htp]
    \centering
\resizebox{\textwidth}{!}{%
\begin{tabular}{|l|l|c|c|c|c|c|c|c|c|c|c|c|} 
\hline
\multicolumn{1}{|c|}{\multirow{2}{*}{\begin{tabular}[c]{@{}c@{}}Polynomial\\Config\end{tabular}}} & \multicolumn{1}{c|}{\multirow{2}{*}{\begin{tabular}[c]{@{}c@{}}Input\\Format\end{tabular}}} & \multicolumn{2}{c|}{Endpoint}                                 & \multirow{2}{*}{\#Train} & \multicolumn{2}{c|}{Full Proof (Beam-1)}                                                                                    & \multicolumn{2}{c|}{Step-wise (Beam-5)}                                                                    & \multicolumn{4}{c|}{Calibration (Beam-5)}                                   \\ 
\cline{3-4}\cline{6-13}
\multicolumn{1}{|c|}{}                                                                            & \multicolumn{1}{c|}{}                                                                       & \#EE & \begin{tabular}[c]{@{}c@{}}Endpoint\\ Acc\end{tabular} &                          & \begin{tabular}[c]{@{}c@{}}Full Proof\\ Accuracy\end{tabular} & \begin{tabular}[c]{@{}c@{}}Stepwise\\ Accuracy\end{tabular} & \begin{tabular}[c]{@{}c@{}}Top-1\\ Acc\end{tabular} & \begin{tabular}[c]{@{}c@{}}Beam-5\\ Acc\end{tabular} & \begin{tabular}[c]{@{}c@{}}Sure\\ Rate\end{tabular} & P     & R     & F1    \\ 
\hline \hline
\multirow{2}{*}{\begin{tabular}[c]{@{}l@{}}Small \\ Coeff\end{tabular}}                           & Infix                                                                                       & 4.3M & 94.7                                                   & 3.7M                     & 87.9                                                          & 97.01                                                       & 88.9                                                & 91                                                   & 81.07                                               & 100   & 91.19 & 0.95  \\ 
\cline{2-13}
                                                                                                  & Prefix                                                                                      & 4.5M & 93.93                                                  & 5.3M                     & \textbf{91.2}                                                 & \textbf{98.08 }                                             & 83.83                                               & 86.7                                                 & 77.57                                               & 100   & 92.52 & 0.96  \\ 
\hline \hline
\multirow{2}{*}{\begin{tabular}[c]{@{}l@{}}Medium\\ Coeff\end{tabular}}                           & Infix                                                                                       & 7M   & 95.3                                                   & 5.3M                     & \textbf{88.5 }                                                & \textbf{97.35}                                              & 90.98                                               & 93.7                                                 & 84.64                                               & 99.98 & 93.01 & 0.96  \\ 
\cline{2-13}
                                                                                                  & Prefix                                                                                      & 5.2M & 92.77                                                  & 4.8M                     & 84.5                                                          & 96.03                                                       & 89.57                                               & 92.93                                                & 81.27                                               & 99.96 & 90.7  & 0.95  \\ 
\hline \hline
\multirow{2}{*}{\begin{tabular}[c]{@{}l@{}}Large \\ Coeff\end{tabular}}                           & Infix                                                                                       & 9M   & 91.8                                                   & 3.8M                     & 80.4                                                          & 95.18                                                       & 90.44                                               & 93.14                                                & 82.74                                               & 99.93 & 91.42 & 0.95  \\ 
\cline{2-13}
                                                                                                  & Prefix                                                                                      & 6.1M & 86.6                                                   & 5.4M                     & \textbf{83.7}                                                 & \textbf{96.23}                                              & 92.23                                               & 94.57                                                & 86.03                                               & 100   & 93.28 & 0.97  \\ 
\hline \hline
\multirow{2}{*}{\begin{tabular}[c]{@{}l@{}}No \\ Backtrack\end{tabular}}                          & Infix                                                                                       & 8.6M & 83.8                                                   & 5M                       & \textbf{72.7 }                                                & \textbf{93.13 }                                             & 75.48                                               & 78.74                                                & 64.4                                                & 100   & 85.32 & 0.92  \\ 
\cline{2-13}
                                                                                                  & Prefix                                                                                      & 7.1M & 79.2                                                   & 4.3M                     & 63.2                                                          & 89.87                                                       & 72.07                                               & 76.43                                                & 59.4                                                & 99.94 & 82.38 & 0.9   \\ 
\hline \hline
\multirow{2}{*}{\begin{tabular}[c]{@{}l@{}}Medium\\ Degree\end{tabular}}                          & Infix                                                                                       & 4.9M & 87.9                                                   & 5.1M                     & 80.5                                                          & 95.13                                                       & 90.3                                                & 92.53                                                & 80.63                                               & 100   & 89.29 & 0.94  \\ 
\cline{2-13}
                                                                                                  & Prefix                                                                                      & 5.2M & 83.73                                                  & 6.1M                     & \textbf{83.4}                                                 & \textbf{96.41}                                              & 92.07                                               & 94.43                                                & 83.13                                               & 99.96 & 90.26 & 0.95  \\ 
\hline \hline
\multirow{2}{*}{\begin{tabular}[c]{@{}l@{}}Medium\\ Terms\end{tabular}}                           & Infix                                                                                       & 8.5M & 90                                                     & 3.8M                     & 64                                                            & 92.03                                                       & 80.5                                                & 83.66                                                & 66.62                                               & 100   & 82.76 & 0.91  \\ 
\cline{2-13}
                                                                                                  & Prefix                                                                                      & 6.6M & 87.07                                                  & 6.3M                     & \textbf{67.8}                                                 & \textbf{93.58}                                              & 89.7                                                & 91.57                                                & 80.33                                               & 99.96 & 89.52 & 0.94  \\
\hline
\end{tabular}%
}
    \caption{Results for 2 variables for the \textsc{Coarse} configuration for both Infix and prefix representations.}
    \label{tab:var2coarse}
\end{table}

\begin{table}[!htpb]
\centering
\resizebox{\textwidth}{!}{%
\begin{tabular}{|l|l|c|c|c|c|c|c|c|c|c|c|c|} 
\hline
\multicolumn{1}{|c|}{\multirow{2}{*}{\begin{tabular}[c]{@{}c@{}}Polynomial\\Config\end{tabular}}} & \multicolumn{1}{c|}{\multirow{2}{*}{\begin{tabular}[c]{@{}c@{}}Proof/Input\\Format\end{tabular}}} & \multicolumn{2}{c|}{Endpoint}                                 & \multirow{2}{*}{\#Train} & \multicolumn{2}{c|}{Full Proof (Beam-1)}                                                                                    & \multicolumn{2}{c|}{Step-wise (Beam-5)}                                                                    & \multicolumn{4}{c|}{Calibration (Beam-5)}                                                     \\ 
\cline{3-4}\cline{6-13}
\multicolumn{1}{|c|}{}                                                                            & \multicolumn{1}{c|}{}                                                                             & \#EE & \begin{tabular}[c]{@{}c@{}}Endpoint\\ Acc\end{tabular} &                          & \begin{tabular}[c]{@{}c@{}}Full Proof\\ Accuracy\end{tabular} & \begin{tabular}[c]{@{}c@{}}Stepwise\\ Accuracy\end{tabular} & \begin{tabular}[c]{@{}c@{}}Top-1\\ Acc\end{tabular} & \begin{tabular}[c]{@{}c@{}}Beam-5\\ Acc\end{tabular} & \begin{tabular}[c]{@{}c@{}}Sure\\ Rate\end{tabular} & P     & R     & F1             \\ \hline \hline

\multirow{2}{*}{{\begin{tabular}[c]{@{}l@{}}SMALL \\ COEFF\end{tabular}}}  & Infix/Fine                                                                      & 4.3M          & 94.7                                                               & 4.6M                              & 88.1                                                                    & 97.19                                                                       & 90.7                                                            & 92.2                                                             & 83.47                                                         & 100        & 92.02      & 0.96        \\ \cline{2-13} 
                                                                                  & Prefix/Fine                                                                     & 4.5M          & 93.93                                                              & 5.4M                              & 90.3                                                                    & 97.83                                                                       & 94.63                                                           & 96.2                                                             & 87.9                                                          & 99.96      & 92.85      & 0.96        \\ \hline \hline
\multirow{2}{*}{{\begin{tabular}[c]{@{}l@{}}MEDIUM \\ COEFF\end{tabular}}} & Infix/Fine                                                                      & 7M            & 95.3                                                               & 4.4M                              & 82.2                                                                    & 96.25                                                                       & 94.28                                                           & 95.76                                                            & 86.24                                                         & 100        & 91.47      & 0.96        \\ \cline{2-13} 
                                                                                  & Prefix/Fine                                                                     & 5.2M          & 92.77                                                              & 2.9M                              & 72.4                                                                    & 93.6                                                                        & 91.53                                                           & 94.33                                                            & 81.97                                                         & 100        & 89.55      & 0.94        \\ \hline \hline
\multirow{2}{*}{{\begin{tabular}[c]{@{}l@{}}LARGE \\ COEFF\end{tabular}}}  & Infix/Fine                                                                      & 9M            & 91.8                                                               & 3.2M                              & 73                                                                      & 93.85                                                                       & 77.94                                                           & 82.2                                                             & 63                                                            & 99.9       & 80.75      & 0.89        \\ \cline{2-13} 
                                                                                  & Prefix/Fine                                                                     & 6.1M          & 86.6                                                               & 4.7M                              & 78.6                                                                    & 95.6                                                                        & 91.93                                                           & 93.47                                                            & 83.87                                                         & 100        & 91.23      & 0.95        \\ \hline \hline
\multirow{2}{*}{{\begin{tabular}[c]{@{}l@{}}NO \\ BACKTRACK\end{tabular}}} & Infix/Fine                                                                      & 8.6M          & 83.8                                                               & 5.8M                              & 72.5                                                                    & 94.64                                                                       & 81.54                                                           & 84.82                                                            & 72.34                                                         & 100        & 88.72      & 0.94        \\ \cline{2-13} 
                                                                                  & Prefix/Fine                                                                     & 7.1M          & 79.2                                                               & 4.1M                              & 60.7                                                                    & 90.48                                                                       & 81.73                                                           & 85.67                                                            & 70.2                                                          & 99.91      & 85.81      & 0.92        \\ \hline \hline
\multirow{2}{*}{{\begin{tabular}[c]{@{}l@{}}MEDIUM \\ DEG\end{tabular}}}   & Infix/Fine                                                                      & 4.9M          & 87.9                                                               & 3.6M                              & 73.5                                                                    & 94.21                                                                       & 89.78                                                           & 92.46                                                            & 77.22                                                         & 100        & 86.01      & 0.92        \\ \cline{2-13} 
                                                                                  & Prefix/Fine                                                                     & 5.2M          & 83.73                                                              & 4.6M                              & 73.6                                                                    & 94.57                                                                       & 86.5                                                            & 89.4                                                             & 76.93                                                         & 100        & 88.94      & 0.94        \\ \hline \hline
\multirow{2}{*}{{\begin{tabular}[c]{@{}l@{}}MEDIUM \\ TERMS\end{tabular}}} & Infix/Fine                                                                      & 8.5M          & 90                                                                 & 4.8M                              & 64                                                                      & 92.98                                                                       & 79.04                                                           & 81.86                                                            & 66.92                                                         & 99.88      & 84.56      & 0.92        \\ \cline{2-13} 
                                                                                  & Prefix/Fine                                                                     & 6.6M          & 87.07                                                              & 4.5M                              & 62.9                                                                    & 92.74                                                                       & 86.4                                                            & 89.07                                                            & 73.67                                                         & 100        & 85.26      & 0.92        \\ \hline
\end{tabular}%
}
\caption{Results for \textsc{fine} configuration for 2 Variables for Infix and Prefix representation (No curriculum, No annotation).}
\label{tab:results2varfine}
\end{table}

\begin{table}[!htpb]
\centering
\resizebox{\textwidth}{!}{%
\begin{tabular}{|c|c|c|c|c|c|c|c|c|c|}
\hline
\multirow{2}{*}{\textbf{Config}}                                                  & \multirow{2}{*}{\textbf{Proof Type}} & \multicolumn{2}{c|}{\textbf{Full Proof}}                                                                                                                   & \multicolumn{6}{c|}{\textbf{Error Percentage}}                                                                                                                                                                                                                                                                                                                                                                             \\ \cline{3-10} 
                                                                                  &                                      & \textbf{\begin{tabular}[c]{@{}c@{}}Full Proof \\  Accuracy\end{tabular}} & \textbf{\begin{tabular}[c]{@{}c@{}}Greedy\\  Stepwise\\  Accuracy\end{tabular}} & \textbf{\begin{tabular}[c]{@{}c@{}}First \\  FacStep\end{tabular}} & \textbf{\begin{tabular}[c]{@{}c@{}}Total \\  FacStep\end{tabular}} & \textbf{\begin{tabular}[c]{@{}c@{}}First \\  MulStep\end{tabular}} & \textbf{\begin{tabular}[c]{@{}c@{}}Total \\  MulStep\end{tabular}} & \textbf{\begin{tabular}[c]{@{}c@{}}First \\  SumStep\end{tabular}} & \textbf{\begin{tabular}[c]{@{}c@{}}Total\\  SumStep\end{tabular}} \\ \hline \hline
\multirow{4}{*}{\textbf{\begin{tabular}[c]{@{}c@{}}SMALL \\ COEFF\end{tabular}}}  & Coarse/Infix                         & 95                                                                       & 98.83                                                                           & 8                                                                  & 9.43                                                               & 88                                                                 & 84.91                                                              & 4                                                                  & 5.66                                                              \\ \cline{2-10} 
                                                                                  & Fine/Infix                           & \textbf{98.9}                                                            & \textbf{99.79}                                                                  & 0                                                                  & 0                                                                  & 100                                                                & 100                                                                & 0                                                                  & 0                                                                 \\ \cline{2-10} 
                                                                                  & Coarse/Prefix                        & 95.3                                                                     & 98.97                                                                           & 4.26                                                               & 4.08                                                               & 72.34                                                              & 71.43                                                              & 23.4                                                               & 24.49                                                             \\ \cline{2-10} 
                                                                                  & Fine/Prefix                          & 96.9                                                                     & 99.4                                                                            & 9.68                                                               & 9.68                                                               & 77.42                                                              & 77.42                                                              & 12.9                                                               & 12.9                                                              \\ \hline \hline
\multirow{4}{*}{\textbf{\begin{tabular}[c]{@{}c@{}}MEDIUM \\ COEFF\end{tabular}}} & Coarse/Infix                         & 92.8                                                                     & 98.24                                                                           & 1.39                                                               & 1.25                                                               & 95.83                                                              & 92.5                                                               & 2.78                                                               & 6.25                                                              \\ \cline{2-10} 
                                                                                  & Fine/Infix                           & 90.3                                                                     & 97.99                                                                           & 11.34                                                              & 11.32                                                              & 85.57                                                              & 84.91                                                              & 3.09                                                               & 3.77                                                              \\ \cline{2-10} 
                                                                                  & Coarse/Prefix                        & \textbf{93.6}                                                            & \textbf{98.58}                                                                  & 3.12                                                               & 2.94                                                               & 95.31                                                              & 95.59                                                              & 1.56                                                               & 1.47                                                              \\ \cline{2-10} 
                                                                                  & Fine/Prefix                          & 91.7                                                                     & 98.37                                                                           & 2.41                                                               & 2.33                                                               & 96.39                                                              & 96.51                                                              & 1.2                                                                & 1.16                                                              \\ \hline \hline
\multirow{4}{*}{\textbf{\begin{tabular}[c]{@{}c@{}}LARGE \\ COEFF\end{tabular}}}  & Coarse/Infix                         & 82.1                                                                     & 95.97                                                                           & 3.35                                                               & 3.02                                                               & 93.85                                                              & 91.46                                                              & 2.79                                                               & 5.53                                                              \\ \cline{2-10} 
                                                                                  & Fine/Infix                           & 82.5                                                                     & 96.44                                                                           & 2.86                                                               & 2.56                                                               & 93.71                                                              & 90.77                                                              & 3.43                                                               & 6.67                                                              \\ \cline{2-10} 
                                                                                  & Coarse/Prefix                        & \textbf{83.5}                                                            & \textbf{96.25}                                                                  & 4.24                                                               & 3.78                                                               & 93.94                                                              & 92.97                                                              & 1.82                                                               & 3.24                                                              \\ \cline{2-10} 
                                                                                  & Fine/Prefix                          & 82                                                                       & 96.32                                                                           & 3.33                                                               & 2.97                                                               & 90.56                                                              & 86.63                                                              & 6.11                                                               & 10.4                                                              \\ \hline \hline
\multirow{4}{*}{\textbf{\begin{tabular}[c]{@{}c@{}}NO \\ BACKTRACK\end{tabular}}} & Coarse/Infix                         & 75.6                                                                     & 94.62                                                                           & 2.87                                                               & 3.13                                                               & 93.44                                                              & 86.83                                                              & 3.69                                                               & 10.03                                                             \\ \cline{2-10} 
                                                                                  & Fine/Infix                           & 74.5                                                                     & 94.76                                                                           & 3.14                                                               & 3.56                                                               & 93.33                                                              & 78.63                                                              & 3.53                                                               & 17.81                                                             \\ \cline{2-10} 
                                                                                  & Coarse/Prefix                        & \textbf{79.7}                                                            & \textbf{95.38}                                                                  & 7.39                                                               & 6.57                                                               & 89.16                                                              & 83.94                                                              & 3.45                                                               & 9.49                                                              \\ \cline{2-10} 
                                                                                  & Fine/Prefix                          & 74.7                                                                     & 95.23                                                                           & 2.37                                                               & 2.41                                                               & 96.44                                                              & 89.16                                                              & 1.19                                                               & 8.43                                                              \\ \hline \hline
\multirow{4}{*}{\textbf{\begin{tabular}[c]{@{}c@{}}MEDIUM \\ DEG\end{tabular}}}   & Coarse/Infix                         & \textbf{92.8}                                                            & \textbf{98.26}                                                                  & 5.56                                                               & 6.02                                                               & 86.11                                                              & 79.52                                                              & 8.33                                                               & 14.46                                                             \\ \cline{2-10} 
                                                                                  & Fine/Infix                           & 83.4                                                                     & 96.12                                                                           & 6.63                                                               & 5.56                                                               & 89.76                                                              & 83.33                                                              & 3.61                                                               & 11.11                                                             \\ \cline{2-10} 
                                                                                  & Coarse/Prefix                        & 87.7                                                                     & 96.82                                                                           & 4.07                                                               & 3.57                                                               & 93.5                                                               & 90.71                                                              & 2.44                                                               & 5.71                                                              \\ \cline{2-10} 
                                                                                  & Fine/Prefix                          & 90.6                                                                     & 97.92                                                                           & 8.51                                                               & 7.55                                                               & 89.36                                                              & 87.74                                                              & 2.13                                                               & 4.72                                                              \\ \hline \hline
\multirow{4}{*}{\textbf{\begin{tabular}[c]{@{}c@{}}MEDIUM \\ TERMS\end{tabular}}} & Coarse/Infix                         & 72.7                                                                     & 93.99                                                                           & 25.64                                                              & 24.18                                                              & 73.26                                                              & 69.72                                                              & 1.1                                                                & 6.1                                                               \\ \cline{2-10} 
                                                                                  & Fine/Infix                           & 75.1                                                                     & 95.42                                                                           & 21.29                                                              & 20.51                                                              & 75.1                                                               & 69.94                                                              & 3.61                                                               & 9.55                                                              \\ \cline{2-10} 
                                                                                  & Coarse/Prefix                        & \textbf{76.3}                                                            & \textbf{95.78}                                                                  & 7.59                                                               & 8.71                                                               & 88.61                                                              & 84.67                                                              & 3.8                                                                & 6.62                                                              \\ \cline{2-10} 
                                                                                  & Fine/Prefix                          & 74.8                                                                     & 95.55                                                                           & 14.68                                                              & 16.76                                                              & 79.76                                                              & 74.28                                                              & 5.56                                                               & 8.96                                                              \\ \hline
\end{tabular}%
}
\caption{Errors for 1 variable in the \textsc{Coarse} and \textsc{Fine} configuration for both Infix and Prefix input representation. (No curriculum, No annotation).}
\label{tab:var1error}
\end{table}

\begin{table}[!htpb]
\centering
\resizebox{\textwidth}{!}{%
\begin{tabular}{|c|c|c|c|c|c|c|c|c|c|}
\hline
\multirow{2}{*}{\textbf{Config}}                                                  & \multirow{2}{*}{\textbf{Proof Type}} & \multicolumn{2}{c|}{\textbf{Full Proof}}                                                                                                                   & \multicolumn{6}{c|}{\textbf{Error Percentage}}                                                                                                                                                                                                                                                                                                                                                                             \\ \cline{3-10} 
                                                                                  &                                      & \textbf{\begin{tabular}[c]{@{}c@{}}Full Proof \\  Accuracy\end{tabular}} & \textbf{\begin{tabular}[c]{@{}c@{}}Greedy\\  Stepwise\\  Accuracy\end{tabular}} & \textbf{\begin{tabular}[c]{@{}c@{}}First \\  FacStep\end{tabular}} & \textbf{\begin{tabular}[c]{@{}c@{}}Total \\  FacStep\end{tabular}} & \textbf{\begin{tabular}[c]{@{}c@{}}First \\  MulStep\end{tabular}} & \textbf{\begin{tabular}[c]{@{}c@{}}Total \\  MulStep\end{tabular}} & \textbf{\begin{tabular}[c]{@{}c@{}}First \\  SumStep\end{tabular}} & \textbf{\begin{tabular}[c]{@{}c@{}}Total\\  SumStep\end{tabular}} \\ \hline \hline
\multirow{4}{*}{\textbf{\begin{tabular}[c]{@{}c@{}}SMALL \\ COEFF\end{tabular}}}  & Coarse/Infix                         & 87.9                                                                     & 97.01                                                                           & 5.79                                                               & 4.9                                                                & 88.43                                                              & 79.72                                                              & 5.79                                                               & 15.38                                                             \\ \cline{2-10} 
                                                                                  & Fine/Infix                           & \textbf{88.1}                                                            & \textbf{97.19}                                                                  & 8.4                                                                & 7.98                                                               & 75.63                                                              & 68.1                                                               & 15.97                                                              & 23.93                                                             \\ \cline{2-10} 
                                                                                  & Coarse/Prefix                        & \textbf{91.2}                                                            & \textbf{98.08}                                                                  & 1.14                                                               & 1.03                                                               & 88.64                                                              & 84.54                                                              & 10.23                                                              & 14.43                                                             \\ \cline{2-10} 
                                                                                  & Fine/Prefix                          & 90.3                                                                     & 97.83                                                                           & 8.25                                                               & 6.35                                                               & 80.41                                                              & 73.02                                                              & 11.34                                                              & 20.63                                                             \\ \hline \hline
\multirow{4}{*}{\textbf{\begin{tabular}[c]{@{}c@{}}MEDIUM \\ COEFF\end{tabular}}} & Coarse/Infix                         & 88.5                                                                     & 97.35                                                                           & 4.35                                                               & 3.73                                                               & 83.48                                                              & 76.87                                                              & 12.17                                                              & 19.4                                                              \\ \cline{2-10} 
                                                                                  & Fine/Infix                           & 82.2                                                                     & 96.25                                                                           & 2.25                                                               & 1.83                                                               & 76.4                                                               & 68.81                                                              & 21.35                                                              & 29.36                                                             \\ \cline{2-10} 
                                                                                  & Coarse/Prefix                        & \textbf{84.5}                                                            & \textbf{96.03}                                                                  & 3.87                                                               & 3.68                                                               & 88.39                                                              & 81.58                                                              & 7.74                                                               & 14.74                                                             \\ \cline{2-10} 
                                                                                  & Fine/Prefix                          & 72.4                                                                     & 93.6                                                                            & 12.68                                                              & 9.95                                                               & 76.09                                                              & 67.74                                                              & 11.23                                                              & 22.31                                                             \\ \hline \hline
\multirow{4}{*}{\textbf{\begin{tabular}[c]{@{}c@{}}LARGE \\ COEFF\end{tabular}}}  & Coarse/Infix                         & 80.4                                                                     & 95.18                                                                           & 6.12                                                               & 4.84                                                               & 82.65                                                              & 75.81                                                              & 11.22                                                              & 19.35                                                             \\ \cline{2-10} 
                                                                                  & Fine/Infix                           & 73                                                                       & 93.85                                                                           & 11.85                                                              & 8.74                                                               & 70.74                                                              & 62.3                                                               & 17.41                                                              & 28.96                                                             \\ \cline{2-10} 
                                                                                  & Coarse/Prefix                        & \textbf{83.7}                                                            & \textbf{96.23}                                                                  & 4.29                                                               & 3.61                                                               & 87.12                                                              & 82.99                                                              & 8.59                                                               & 13.4                                                              \\ \cline{2-10} 
                                                                                  & Fine/Prefix                          & 78.6                                                                     & 95.6                                                                            & 5.14                                                               & 4.2                                                                & 81.31                                                              & 74.43                                                              & 13.55                                                              & 21.37                                                             \\ \hline \hline
\multirow{4}{*}{\textbf{\begin{tabular}[c]{@{}c@{}}NO \\ BACKTRACK\end{tabular}}} & Coarse/Infix                         & \textbf{72.7}                                                            & \textbf{93.13}                                                                  & 4.4                                                                & 3.15                                                               & 87.55                                                              & 75.79                                                              & 8.06                                                               & 21.07                                                             \\ \cline{2-10} 
                                                                                  & Fine/Infix                           & 72.5                                                                     & 94.64                                                                           & 3.27                                                               & 2.54                                                               & 85.09                                                              & 73.79                                                              & 11.64                                                              & 23.66                                                             \\ \cline{2-10} 
                                                                                  & Coarse/Prefix                        & \textbf{63.2}                                                            & \textbf{89.87}                                                                  & 3.26                                                               & 2.24                                                               & 91.3                                                               & 78.73                                                              & 5.43                                                               & 19.03                                                             \\ \cline{2-10} 
                                                                                  & Fine/Prefix                          & 60.7                                                                     & 90.48                                                                           & 2.29                                                               & 1.58                                                               & 89.31                                                              & 72.64                                                              & 8.4                                                                & 25.79                                                             \\ \hline \hline
\multirow{4}{*}{\textbf{\begin{tabular}[c]{@{}c@{}}MEDIUM \\ DEG\end{tabular}}}   & Coarse/Infix                         & \textbf{80.5}                                                            & \textbf{95.13}                                                                  & 6.67                                                               & 5.44                                                               & 81.54                                                              & 71.97                                                              & 11.79                                                              & 22.59                                                             \\ \cline{2-10} 
                                                                                  & Fine/Infix                           & 73.5                                                                     & 94.21                                                                           & 7.17                                                               & 6.55                                                               & 68.3                                                               & 57.83                                                              & 24.53                                                              & 35.61                                                             \\ \cline{2-10} 
                                                                                  & Coarse/Prefix                        & \textbf{83.4}                                                            & \textbf{96.41}                                                                  & 4.82                                                               & 4.19                                                               & 81.33                                                              & 75.39                                                              & 13.86                                                              & 20.42                                                             \\ \cline{2-10} 
                                                                                  & Fine/Prefix                          & 73.6                                                                     & 94.57                                                                           & 7.58                                                               & 6.38                                                               & 75.38                                                              & 67.48                                                              & 17.05                                                              & 26.14                                                             \\ \hline \hline
\multirow{4}{*}{\textbf{\begin{tabular}[c]{@{}c@{}}MEDIUM \\ TERMS\end{tabular}}} & Coarse/Infix                         & 64                                                                       & 92.03                                                                           & 25                                                                 & 19.05                                                              & 72.5                                                               & 66.5                                                               & 2.5                                                                & 14.45                                                             \\ \cline{2-10} 
                                                                                  & Fine/Infix                           & 64                                                                       & 92.98                                                                           & 13.61                                                              & 8.62                                                               & 79.44                                                              & 69.59                                                              & 6.94                                                               & 21.79                                                             \\ \cline{2-10} 
                                                                                  & Coarse/Prefix                        & \textbf{67.8}                                                            & \textbf{93.58}                                                                  & 10.25                                                              & 7.69                                                               & 87.89                                                              & 80.98                                                              & 1.86                                                               & 11.32                                                             \\ \cline{2-10} 
                                                                                  & Fine/Prefix                          & 62.9                                                                     & 92.74                                                                           & 9.16                                                               & 5.97                                                               & 85.71                                                              & 74.37                                                              & 5.12                                                               & 19.65                                                             \\ \hline
\end{tabular}%
}
\caption{Errors for 2 variables in the \textsc{Coarse} and \textsc{Fine} configuration for both Infix and Prefix input representation. (No curriculum, No annotation).}
\label{tab:var2error}
\end{table}

\section{Symbolic Proof and Hyperparameter Tuning (Additional Results)}
\label{sec:appendx:calc}
We present the step-wise error analysis of the best models from hyper-parameter tuning experiment in Table \ref{tab:var1comboerrors}. For comparison, we also include the errors made by the best models in the symbolic calculator setting for corresponding configurations. 

\paragraph{Hyperparameter Tuning Observations} 
$\bullet$ For learning rates (\texttt{lr}), we first started with the range $0.01, 0.005, 0.0005$ and experimented with few configurations, \textsc{Small Coeff} and \textsc{Medium Terms}. \texttt{lr} greater than $0.0005$ resulted in zero validation scores. Later, in our experiments, we settled with the range $0.0001, 0.0005, 0.00001$. Similarly, \texttt{lr} $10^{-5}$ took a long time to converge and validation score started oscillating. Dropout choices did not show any particular advantage over the other. For most cases, configuration with batch size 64 showed dominant results. \\
$\bullet$  Apart from a limited number of settings, \textsc{coarse} proof granularity resulted in the best model. This is expected as \textsc{fine} proof creates long proofs. \\
$\bullet$ Contrary to the observation made in the \textsc{Transformers-S} configuration, infix representation consistently improved over prefix. This is observed for the symbolic calculator setting as well.

\begin{table}[!htb]
\centering
\resizebox{\textwidth}{!}{%
\begin{tabular}{|c|l|l|l|l|l|l|l|l|l|l|}
\hline
                                                                         &            & \begin{tabular}[c]{@{}l@{}}Model\\ Info\end{tabular}                                & \begin{tabular}[c]{@{}l@{}}Full Proof \\ Acc\end{tabular} & \begin{tabular}[c]{@{}l@{}}Stepwise\\ Acc\end{tabular} & \begin{tabular}[c]{@{}l@{}}First\\ FacSimp\end{tabular} & \begin{tabular}[c]{@{}l@{}}Total\\ FacSimp\end{tabular} & \begin{tabular}[c]{@{}l@{}}First\\ MulSimp\end{tabular} & \begin{tabular}[c]{@{}l@{}}Total\\ MulSimp\end{tabular} & \begin{tabular}[c]{@{}l@{}}First\\ SumSimp\end{tabular} & \begin{tabular}[c]{@{}l@{}}Total\\ SumSimp\end{tabular} \\ \hline
\multirow{4}{*}{\begin{tabular}[c]{@{}c@{}}Large\\ Coeff\end{tabular}}   & h/prefix   & \begin{tabular}[c]{@{}l@{}}Emb: 512,\\ d/o: 0.5, bs: 64\\ Tr-L, Coarse\end{tabular} & 78.8                                                      & 95.0                                                   & 8.96                                                    & 9.4                                                     & 90.09                                                   & 88.03                                                   & 0.94                                                    & 2.56                                                    \\ \cline{2-11} 
                                                                         & h/infix    & \begin{tabular}[c]{@{}l@{}}Emb: 256,\\ d/o: 0, bs: 32\\ Tr-L, Fine\end{tabular}     & 84.6                                                      & 96.79                                                  & 5.19                                                    & 5.68                                                    & 94.81                                                   & 92.05                                                   & 0.00                                                    & 2.27                                                    \\ \cline{2-11} 
 & \texttt{cal}/prefix & \begin{tabular}[c]{@{}l@{}}Emb: 512,\\ d/o: 0.5, bs: 64\\ Tr-L, Coarse\end{tabular} & \textbf{94.6} & \textbf{98.98} & 18.52  & 18.97  & 79.63  & 79.31 & 1.85 & 1.72                                                    \\ \cline{2-11} 
  & \texttt{cal}/infix  & \begin{tabular}[c]{@{}l@{}}Emb: 512,\\ d/o: 0.5, bs: 64\\ Tr-L, Coarse\end{tabular} & \textbf{95.7}                & \textbf{99.16}  & 4.65  & 6.25  & 83.72   & 79.17  & 11.63  & 14.58 \\ \hline
\multirow{4}{*}{\begin{tabular}[c]{@{}c@{}}No\\ Backtrack\end{tabular}}  & h/prefix   & \begin{tabular}[c]{@{}l@{}}Emb: 256,\\ d/o: 0, bs: 64\\ Tr-L, Coarse\end{tabular}   & 74.1                                                      & 93.69                                                  & 1.93                                                    & 1.52                                                    & 96.91                                                   & 87.54                                                   & 1.16                                                    & 10.94                                                   \\ \cline{2-11} 
                                                                         & h/infix    & \begin{tabular}[c]{@{}l@{}}Emb: 512,\\ d/o: 0, bs: 64\\ Tr-L, Fine\end{tabular}     & 82.6                                                      & 96.85                                                  & 5.17                                                    & 6.85                                                    & 91.95                                                   & 84.47                                                   & 2.87                                                    & 8.68                                                    \\ \cline{2-11} 
                                                                         & \texttt{cal}/prefix & \begin{tabular}[c]{@{}l@{}}Emb: 512,\\ d/o: 0.5, bs: 32\\ Tr-L, Fine\end{tabular}   & \textbf{83.5}                                                      & \textbf{97.25}                                                  & 11.52                                                   & 13.24                                                   & 86.67                                                   & 81.74                                                   & 1.82                                                    & 5.02                                                    \\ \cline{2-11} 
\cline{2-11}  & \texttt{cal}/infix**  & \begin{tabular}[c]{@{}l@{}}Emb: 512,\\ d/o: 0.5, bs: 64\\ Tr-S, Fine\end{tabular}   & \textbf{87.2}                                                      & \textbf{97.94}                                                  & 7.81                                                    & 10.98                                                   & 86.72                                                   & 80.49                                                   & 5.47                                                    & 8.54                                                    \\ \hline \hline
\multirow{4}{*}{\begin{tabular}[c]{@{}c@{}}Medium\\ Degree\end{tabular}} & h/prefix   & \begin{tabular}[c]{@{}l@{}}Emb: 256,\\ d/o: 0.5, bs: 32\\ Tr-S, Coarse\end{tabular} & 82                                                        & 95.49                                                  & 11.11                                                   & 10.05                                                   & 85.56                                                   & 84.42                                                   & 3.33                                                    & 5.53                                                    \\ \cline{2-11} 
                                                                         & h/infix    & \begin{tabular}[c]{@{}l@{}}Emb: 512,\\ d/o: 0, bs: 64\\ Tr-L, Fine\end{tabular}     & 90.8                                                      & 97.81                                                  & 2.17                                                    & 1.79                                                    & 90.22                                                   & 83.93                                                   & 7.61                                                    & 14.29                                                   \\ \cline{2-11} 
                                                                         & \texttt{cal}/prefix & \begin{tabular}[c]{@{}l@{}}Emb: 512,\\ d/o: 0.5, bs: 64\\ Tr-L, Coarse\end{tabular} & 90.5                                                      & 97.89                                                  & 10.53                                                   & 9.65                                                    & 83.16                                                   & 78.95                                                   & 6.32                                                    & 11.40                                                   \\ \cline{2-11} 
    & \texttt{cal}/infix  & \begin{tabular}[c]{@{}l@{}}Emb: 512,\\ d/o: 0.5, bs: 64\\ Tr-L, Coarse\end{tabular} & \textbf{90.5}                                             & \textbf{98.08}                                         & 5.26                                                    & 4.81                                                    & 91.58                                                   & 87.5                                                    & 3.16                                                    & 7.69                                                    \\ \hline
\multirow{4}{*}{\begin{tabular}[c]{@{}c@{}}Medium\\ Terms\end{tabular}}  & h/prefix   & \begin{tabular}[c]{@{}l@{}}Emb: 512,\\ d/o: 0, bs: 128\\ Tr-L, Fine\end{tabular}    & 76.5                                                      & 95.59                                                  & 20.0                                                    & 22.45                                                   & 71.49                                                   & 66.47                                                   & 8.51                                                    & 11.08                                                   \\ \cline{2-11} 
                                                                         & h/infix    & \begin{tabular}[c]{@{}l@{}}Emb: 512,\\ d/o: 0, bs: 32\\ Tr-S, Fine\end{tabular}     & 79.5                                                      & 96.49                                                  & 18.54                                                   & 17.58                                                   & 80.49                                                   & 77.29                                                   & 0.98                                                    & 5.13                                                    \\ \cline{2-11} 
                                                                         & \texttt{cal}/prefix & \begin{tabular}[c]{@{}l@{}}Emb: 512,\\ d/o: 0.5, bs: 32\\ Tr-L, Coarse\end{tabular} & 88.3                                                      & 98.35                                                  & 9.4                                                     & 10.85                                                   & 88.03                                                   & 86.82                                                   & 2.56                                                    & 2.33                                                    \\ \cline{2-11} 
                                                                         & \texttt{cal}/infix  & \begin{tabular}[c]{@{}l@{}}Emb: 256,\\ d/o: 0.5, bs: 128\\ Tr-L, Fine\end{tabular}  & \textbf{89.5}                                             & \textbf{98.45}                                         & 20.0                                                    & 25.00                                                   & 74.29                                                   & 66.91                                                   & 5.71                                                    & 8.09                                                    \\ \hline
\end{tabular}%
}
\caption{Error analysis for 1 variable, hyperparmeter-tuning and symbolic calculator experiments. \textit{h/prefix} and \textit{h/infix} rows denote the error analysis of models with best validation scores after the hyper-parameter search for prefix and infix input representation respectively. Similarly for \textit{cal/prefix} and \textit{cal/infix} provides the error analysis for the calculator setting for prefix and infix for the best models after hyperparameter search. For each row, we also show the best model and hyper-parameter configuration. For all configurations, learning rate $10^{-4}$ produced the best validation scores.  **Some runs for the \texttt{cal}/infix \textsc{No Backtrack} has not finished. We will update the final numbers during publication. }
\label{tab:var1comboerrors}
\end{table}

\section{Annotated Proof}
\label{sec:appendx:ann}
In each step, simplification is performed over a sub-expression of the polynomial. To check explicitly, if the system can locate the sub-expression and find the type of simplification step, we devise the annotated proof setting.  For each simplification step, we add an intermediate step, in which the model \textit{annotates} the part of polynomial to operate on. For example, the starting input sequence is ``$\text{MARK} ~\$~ (5*x_1^2 + x_1*x_2)*(3*x_1)*(2)$''; and the corresponding expected output sequence is ``$\text{MUL} ~\$~ \#(5*x_1^2 + x_1*x_2)*(3*x_1)\#*(2)$''. Each sequence has two parts: 1) the step index to perform ({\sc Mark, Mul, Fac, Sum}), and 2) the polynomial expression. For \textsc{Mark} step, a marker token (\#) is used to annotate the candidate sub-expression to be simplified next.

We experiment only with \textsc{infix} representation. The results for 1 variable and 2 variables are in Table \ref{tab:annotate1var} and \ref{tab:annotate2var}. The errors per step type are shown in  Tables \ref{tab:annotate1varerror} and \ref{tab:annotate2varerror}. Compared to non-annotated setting, while the step-wise accuracy is similar, the proof accuracy suffers often by 7-10\%. A reason for such decrease in accuracy is that length of the annotated proofs are twice as long as non-annotated. However, we note that the  errors in \textsc{Mark} step are the lowest compared to other types of steps. This indicates that the models are able to learn the candidate sub-expression for simplification, and predict the next operation correctly.

\begin{table}[!htpb]
\centering
\resizebox{\textwidth}{!}{%
\begin{tabular}{|l|l|c|c|c|c|c|c|c|c|c|c|c|} 
\hline
\multicolumn{1}{|c|}{\multirow{2}{*}{\begin{tabular}[c]{@{}c@{}}Polynomial\\Config\end{tabular}}} & \multicolumn{1}{c|}{\multirow{2}{*}{\begin{tabular}[c]{@{}c@{}}Proof/Input\\Format\end{tabular}}} & \multicolumn{2}{c|}{Endpoint}                                 & \multirow{2}{*}{\#Train} & \multicolumn{2}{c|}{Full Proof (Beam-1)}                                                                                    & \multicolumn{2}{c|}{Step-wise (Beam-5)}                                                                    & \multicolumn{4}{c|}{Calibration (Beam-5)}                                                     \\ 
\cline{3-4}\cline{6-13}
\multicolumn{1}{|c|}{}                                                                            & \multicolumn{1}{c|}{}                                                                             & \#EE & \begin{tabular}[c]{@{}c@{}}Endpoint\\ Acc\end{tabular} &                          & \begin{tabular}[c]{@{}c@{}}Full Proof\\ Accuracy\end{tabular} & \begin{tabular}[c]{@{}c@{}}Stepwise\\ Accuracy\end{tabular} & \begin{tabular}[c]{@{}c@{}}Top-1\\ Acc\end{tabular} & \begin{tabular}[c]{@{}c@{}}Beam-5\\ Acc\end{tabular} & \begin{tabular}[c]{@{}c@{}}Sure\\ Rate\end{tabular} & P     & R     & F1             \\ \hline
\multirow{2}{*}{{\begin{tabular}[c]{@{}c@{}}SMALL\\COEFF\end{tabular}}}  & Fine                                 & \multirow{2}{*}{5M}                                                       & \multirow{2}{*}{96}                                                    & 2.4M                                                                                    & 88.5                                                                     & 98.82                                                                           & 86.77                                                               & 87.7                                                                 & 83.97              & 99.96      & 96.73      & 0.98        \\ \cline{2-2} \cline{5-13} 
                                       & Coarse                               &                                                                           &                                                                        & 3.7M                                                                                    & \textbf{91.9}                                                            & \textbf{99.16}                                                                  & 90.07                                                               & 90.73                                                                & 88.13              & 100        & 97.85      & 0.99        \\ \hline
\multirow{2}{*}{{\begin{tabular}[c]{@{}c@{}}MEDIUM\\COEFF\end{tabular}}} & Fine                                 & \multirow{2}{*}{4.1M}                                                     & \multirow{2}{*}{91.2}                                                  & 2.8M                                                                                    & 78.6                                                                     & 97.66                                                                           & 92.67                                                               & 93.63                                                                & 88.23              & 100        & 95.22      & 0.98        \\ \cline{2-2} \cline{5-13} 
                                       & Coarse                               &                                                                           &                                                                        & 3.5M                                                                                    & \textbf{84.2}                                                            & \textbf{98.29}                                                                  & 94.83                                                               & 95.53                                                                & 92.4               & 99.96      & 97.4       & 0.99        \\ \hline
\multirow{2}{*}{{\begin{tabular}[c]{@{}c@{}}LARGE\\COEFF\end{tabular}}}  & Fine                                 & \multirow{2}{*}{4.8M}                                                     & \multirow{2}{*}{83.73}                                                 & 3.6M                                                                                    & 75.5                                                                     & 97.37                                                                           & 96.8                                                                & 97.8                                                                 & 92.4               & 99.93      & 95.39      & 0.98        \\ \cline{2-2} \cline{5-13} 
                                       & Coarse                               &                                                                           &                                                                        & 4.6M                                                                                    & \textbf{80.3}                                                            & \textbf{97.86}                                                                  & 80.37                                                               & 81.6                                                                 & 77.83              & 100        & 96.85      & 0.98        \\ \hline
\multirow{2}{*}{{\begin{tabular}[c]{@{}c@{}}NO BACK\\TRACK\end{tabular}}} & Fine                                 & \multirow{2}{*}{5.9M}                                                     & \multirow{2}{*}{80.1}                                                  & 4.1M                                                                                    & \textbf{68}                                                              & \textbf{96.78}                                                                  & 90.43                                                               & 92.33                                                                & 84.5               & 99.96      & 93.4       & 0.97        \\ \cline{2-2} \cline{5-13} 
                                       & Coarse                               &                                                                           &                                                                        & 3.6M                                                                                    & 59.7                                                                     & 95                                                                              & 92.5                                                                & 94.33                                                                & 86.47              & 99.81      & 93.3       & 0.96        \\ \hline
\multirow{2}{*}{{\begin{tabular}[c]{@{}c@{}}MEDIUM\\DEG\end{tabular}}}   & Fine                                 & \multirow{2}{*}{9.2M}                                                     & \multirow{2}{*}{96.4}                                                  & 3.7M                                                                                    & 76                                                                       & 97.37                                                                           & 83.67                                                               & 85.23                                                                & 79.1               & 100        & 94.54      & 0.97        \\ \cline{2-2} \cline{5-13} 
                                       & Coarse                               &                                                                           &                                                                        & 3.4M                                                                                    & \textbf{78.7}                                                            & \textbf{97.38}                                                                  & 93.2                                                                & 94.37                                                                & 88.2               & 100        & 94.64      & 0.97        \\ \hline
\multirow{2}{*}{{\begin{tabular}[c]{@{}c@{}}MEDIUM\\TERMS\end{tabular}}} & Fine                                 & \multirow{2}{*}{4.6M}                                                     & \multirow{2}{*}{81.9}                                                  & 3.6M                                                                                    & \textbf{70.4}                                                            & \textbf{97.48}                                                                  & 91.5                                                                & 92.2                                                                 & 86.87              & 100        & 94.94      & 0.97        \\ \cline{2-2} \cline{5-13} 
                                       & Coarse                               &                                                                           &                                                                        & 3.3M                                                                                    & 66.2                                                                     & 96.34                                                                           & 88.9                                                                & 90.27                                                                & 83.17              & 99.84      & 93.4       & 0.97        \\ \hline
\end{tabular}%
}
\caption{Results for \textsc{fine} and  \textsc{coarse} configurations for 1 Variable for annotated proofs}
\label{tab:annotate1var}
\end{table}

\begin{table}[!htpb]
\centering
\resizebox{\textwidth}{!}{%
\begin{tabular}{|l|l|c|c|c|c|c|c|c|c|c|c|c|} 
\hline
\multicolumn{1}{|c|}{\multirow{2}{*}{\begin{tabular}[c]{@{}c@{}}Polynomial\\Config\end{tabular}}} & \multicolumn{1}{c|}{\multirow{2}{*}{\begin{tabular}[c]{@{}c@{}}Proof/Input\\Format\end{tabular}}} & \multicolumn{2}{c|}{Endpoint}                                 & \multirow{2}{*}{\#Train} & \multicolumn{2}{c|}{Full Proof (Beam-1)}                                                                                    & \multicolumn{2}{c|}{Step-wise (Beam-5)}                                                                    & \multicolumn{4}{c|}{Calibration (Beam-5)}                                                     \\ 
\cline{3-4}\cline{6-13}
\multicolumn{1}{|c|}{}                                                                            & \multicolumn{1}{c|}{}                                                                             & \#EE & \begin{tabular}[c]{@{}c@{}}Endpoint\\ Acc\end{tabular} &                          & \begin{tabular}[c]{@{}c@{}}Full Proof\\ Accuracy\end{tabular} & \begin{tabular}[c]{@{}c@{}}Stepwise\\ Accuracy\end{tabular} & \begin{tabular}[c]{@{}c@{}}Top-1\\ Acc\end{tabular} & \begin{tabular}[c]{@{}c@{}}Beam-5\\ Acc\end{tabular} & \begin{tabular}[c]{@{}c@{}}Sure\\ Rate\end{tabular} & P     & R     & F1             \\ \hline \hline
\multirow{2}{*}{\textbf{SMALL COEFF}}  & Fine                                 & \multirow{2}{*}{4.3M}                                                     & \multirow{2}{*}{94.7}                                                  & 3.6M                                                                                    & 82.3                                                                     & 97.93                                                                           & 86.47                                                               & 87.5                                                                 & 81.83              & 100        & 94.64      & 0.97        \\ \cline{2-2} \cline{5-13} 
                                       & Coarse                               &                                                                           &                                                                        & 5.1M                                                                                    & \textbf{85}                                                              & \textbf{98.31}                                                                  & 93.5                                                                & 94.03                                                                & 90.27              & 100        & 96.54      & 0.98        \\ \hline \hline
\multirow{2}{*}{\textbf{MEDIUM COEFF}} & Fine                                 & \multirow{2}{*}{7M}                                                       & \multirow{2}{*}{95.3}                                                  & 5.4M                                                                                    & 78.8                                                                     & 97.78                                                                           & 93.8                                                                & 94.5                                                                 & 90.2               & 99.93      & 96.09      & 0.98        \\ \cline{2-2} \cline{5-13} 
                                       & Coarse                               &                                                                           &                                                                        & 5M                                                                                      & \textbf{80.1}                                                            & \textbf{97.69}                                                                  & 89.37                                                               & 90.27                                                                & 86.77              & 99.96      & 97.05      & 0.98        \\ \hline \hline
\multirow{2}{*}{\textbf{LARGE COEFF}}  & Fine                                 & \multirow{2}{*}{9M}                                                       & \multirow{2}{*}{91.8}                                                  & 4.1M                                                                                    & 70.1                                                                     & 96.59                                                                           & 84.8                                                                & 86.63                                                                & 77.77              & 99.83      & 91.55      & 0.96        \\ \cline{2-2} \cline{5-13} 
                                       & Coarse                               &                                                                           &                                                                        & 4M                                                                                      & \textbf{73.2}                                                            & \textbf{96.66}                                                                  & 92.77                                                               & 93.8                                                                 & 87.23              & 100        & 94.04      & 0.97        \\ \hline \hline
\multirow{2}{*}{\textbf{NO BACKTRACK}} & Fine                                 & \multirow{2}{*}{8.6M}                                                     & \multirow{2}{*}{83.8}                                                  & 3.5M                                                                                    & \textbf{46.5}                                                            & \textbf{92.93}                                                                  & 84.9                                                                & 87.67                                                                & 74.5               & 99.96      & 87.71      & 0.93        \\ \cline{2-2} \cline{5-13} 
                                       & Coarse                               &                                                                           &                                                                        & 6.7M                                                                                    & \textbf{65.5}                                                            & \textbf{95.7}                                                                   & 67.8                                                                & 69.37                                                                & 63.3               & 99.79      & 93.17      & 0.96        \\ \hline \hline
\multirow{2}{*}{\textbf{MEDIUM DEG}}   & Fine                                 & \multirow{2}{*}{4.9M}                                                     & \multirow{2}{*}{87.9}                                                  & 3.9M                                                                                    & 59.6                                                                     & 95.28                                                                           & 94.13                                                               & 95.7                                                                 & 86.4               & 100        & 91.78      & 0.96        \\ \cline{2-2} \cline{5-13} 
                                       & Coarse                               &                                                                           &                                                                        & 4.1M                                                                                    & \textbf{65.1}                                                            & \textbf{95.61}                                                                  & 85.43                                                               & 87.43                                                                & 78.2               & 99.96      & 91.49      & 0.96        \\ \hline \hline
\multirow{2}{*}{\textbf{MEDIUM TERMS}} & Fine                                 & \multirow{2}{*}{8.5M}                                                     & \multirow{2}{*}{90}                                                    & 4.8M                                                                                    & \textbf{56.9}                                                            & \textbf{95.7}                                                                   & 92.4                                                                & 93.83                                                                & 85.77              & 99.88      & 92.71      & 0.96        \\ \cline{2-2} \cline{5-13} 
                                       & Coarse                               &                                                                           &                                                                        & 4.2M                                                                                    & 52.8                                                                     & 94.57                                                                           & 84                                                                  & 85.93                                                                & 75.93              & 99.82      & 90.24      & 0.95        \\ \hline
\end{tabular}%
}
\caption{Results for \textsc{fine} and  \textsc{coarse} configurations for 2 Variables for annotated proofs (No curriculum).}
\label{tab:annotate2var}
\end{table}

\begin{table}[!htpb]
\centering
\resizebox{\textwidth}{!}{%
\begin{tabular}{|c|c|c|c|c|c|c|c|c|c|c|c|}
\hline
\multirow{2}{*}{\textbf{Config}}       & \multirow{2}{*}{\textbf{Proof Type}} & \multicolumn{2}{c|}{\textbf{Full Proof}}                                                                                                                   & \multicolumn{8}{c|}{\textbf{Error Percentage}}                                                                                                                                                                                                                                                                                                                                                                                                                                                                                                                        \\ \cline{3-12} 
                                       &                                      & \textbf{\begin{tabular}[c]{@{}c@{}}Full Proof \\  Accuracy\end{tabular}} & \textbf{\begin{tabular}[c]{@{}c@{}}Greedy\\  Stepwise\\  Accuracy\end{tabular}} & \textbf{\begin{tabular}[c]{@{}c@{}}First \\  FacStep\end{tabular}} & \textbf{\begin{tabular}[c]{@{}c@{}}Total \\  FacStep\end{tabular}} & \textbf{\begin{tabular}[c]{@{}c@{}}First \\  MulStep\end{tabular}} & \textbf{\begin{tabular}[c]{@{}c@{}}Total \\  MulStep\end{tabular}} & \textbf{\begin{tabular}[c]{@{}c@{}}First \\  SumStep\end{tabular}} & \textbf{\begin{tabular}[c]{@{}c@{}}Total\\  SumStep\end{tabular}} & \textbf{\begin{tabular}[c]{@{}c@{}}First \\  MarkStep\end{tabular}} & \textbf{\begin{tabular}[c]{@{}c@{}}Total\\  MarkStep\end{tabular}} \\ \hline \hline
\multirow{2}{*}{\textbf{SMALL COEFF}}  & Fine                                 & 88.5                                                                     & 98.82                                                                           & 3.48                                                               & 2.99                                                               & 89.57                                                              & 83.58                                                              & 6.09                                                               & 11.19                                                             & 0.87                                                                & 2.24                                                               \\ \cline{2-12} 
                                       & Coarse                               & \textbf{91.9}                                                            & \textbf{99.16}                                                                  & 1.23                                                               & 1.19                                                               & 98.77                                                              & 96.43                                                              & 0                                                                  & 1.19                                                              & 0                                                                   & 1.19                                                               \\ \hline \hline
\multirow{2}{*}{\textbf{MEDIUM COEFF}} & Fine                                 & 78.6                                                                     & 97.66                                                                           & 18.69                                                              & 15.19                                                              & 74.77                                                              & 74.44                                                              & 3.27                                                               & 6.67                                                              & 3.27                                                                & 3.7                                                                \\ \cline{2-12} 
                                       & Coarse                               & \textbf{84.2}                                                            & \textbf{98.29}                                                                  & 4.43                                                               & 4.65                                                               & 84.81                                                              & 84.88                                                              & 6.33                                                               & 5.81                                                              & 4.43                                                                & 4.65                                                               \\ \hline \hline
\multirow{2}{*}{\textbf{LARGE COEFF}}  & Fine                                 & 75.5                                                                     & 97.37                                                                           & 11.43                                                              & 9.21                                                               & 72.65                                                              & 66.35                                                              & 10.61                                                              & 19.68                                                             & 5.31                                                                & 4.76                                                               \\ \cline{2-12} 
                                       & Coarse                               & \textbf{80.3}                                                            & \textbf{97.86}                                                                  & 5.58                                                               & 5.86                                                               & 90.86                                                              & 87.39                                                              & 1.02                                                               & 4.5                                                               & 2.54                                                                & 2.25                                                               \\ \hline \hline
\multirow{2}{*}{\textbf{NO BACKTRACK}} & Fine                                 & \textbf{68}                                                              & \textbf{96.78}                                                                  & 7.19                                                               & 6.46                                                               & 86.56                                                              & 78.54                                                              & 5.62                                                               & 12.71                                                             & 0.62                                                                & 2.29                                                               \\ \cline{2-12} 
                                       & Coarse                               & \textbf{59.7}                                                            & \textbf{95}                                                                     & 6.2                                                                & 5.25                                                               & 88.09                                                              & 76.88                                                              & 3.72                                                               & 15.41                                                             & 1.99                                                                & 2.45                                                               \\ \hline \hline
\multirow{2}{*}{\textbf{MEDIUM DEG}}   & Fine                                 & 76                                                                       & 97.37                                                                           & 11.67                                                              & 10.85                                                              & 82.5                                                               & 80.34                                                              & 3.75                                                               & 6.78                                                              & 2.08                                                                & 2.03                                                               \\ \cline{2-12} 
                                       & Coarse                               & \textbf{78.7}                                                            & \textbf{97.38}                                                                  & 6.1                                                                & 5.84                                                               & 86.38                                                              & 81.32                                                              & 4.69                                                               & 9.73                                                              & 2.82                                                                & 3.11                                                               \\ \hline \hline
\multirow{2}{*}{\textbf{MEDIUM TERMS}} & Fine                                 & \textbf{70.4}                                                            & \textbf{97.48}                                                                  & 16.89                                                              & 16.27                                                              & 75                                                                 & 69.14                                                              & 3.72                                                               & 8.85                                                              & 4.39                                                                & 5.74                                                               \\ \cline{2-12} 
                                       & Coarse                               & 66.2                                                                     & 96.34                                                                           & 25.44                                                              & 25.28                                                              & 68.05                                                              & 63.48                                                              & 2.37                                                               & 5.81                                                              & 4.14                                                                & 5.43                                                               \\ \hline
\end{tabular}%
}
\caption{Errors for \textsc{fine} and  \textsc{coarse} configurations for 1 Variable for annotated proofs (No curriculum).}
\label{tab:annotate1varerror}
\end{table}

\begin{table}[!htpb]
\centering
\resizebox{\textwidth}{!}{%
\begin{tabular}{|c|c|c|c|c|c|c|c|c|c|c|c|}
\hline
\multirow{2}{*}{\textbf{Config}}       & \multirow{2}{*}{\textbf{Proof Type}} & \multicolumn{2}{c|}{\textbf{Full Proof}}                                                                                                                   & \multicolumn{8}{c|}{\textbf{Error Percentage}}                                                                                                                                                                                                                                                                                                                                                                                                                                                                                                                        \\ \cline{3-12} 
                                       &                                      & \textbf{\begin{tabular}[c]{@{}c@{}}Full Proof \\  Accuracy\end{tabular}} & \textbf{\begin{tabular}[c]{@{}c@{}}Greedy\\  Stepwise\\  Accuracy\end{tabular}} & \textbf{\begin{tabular}[c]{@{}c@{}}First \\  FacStep\end{tabular}} & \textbf{\begin{tabular}[c]{@{}c@{}}Total \\  FacStep\end{tabular}} & \textbf{\begin{tabular}[c]{@{}c@{}}First \\  MulStep\end{tabular}} & \textbf{\begin{tabular}[c]{@{}c@{}}Total \\  MulStep\end{tabular}} & \textbf{\begin{tabular}[c]{@{}c@{}}First \\  SumStep\end{tabular}} & \textbf{\begin{tabular}[c]{@{}c@{}}Total\\  SumStep\end{tabular}} & \textbf{\begin{tabular}[c]{@{}c@{}}First \\  MarkStep\end{tabular}} & \textbf{\begin{tabular}[c]{@{}c@{}}Total\\  MarkStep\end{tabular}} \\ \hline \hline
\multirow{2}{*}{\textbf{SMALL COEFF}}  & Fine                                 & 82.3                                                                     & 97.93                                                                           & 4.52                                                               & 3.07                                                               & 86.44                                                              & 68.97                                                              & 7.34                                                               & 24.14                                                             & 1.69                                                                & 3.83                                                               \\ \cline{2-12} 
                                       & Coarse                               & \textbf{85}                                                              & \textbf{98.31}                                                                  & 2                                                                  & 1.68                                                               & 88.67                                                              & 78.21                                                              & 8                                                                  & 18.44                                                             & 1.33                                                                & 1.68                                                               \\ \hline \hline
\multirow{2}{*}{\textbf{MEDIUM COEFF}} & Fine                                 & 78.8                                                                     & 97.78                                                                           & 8.96                                                               & 6.79                                                               & 80.19                                                              & 68.21                                                              & 9.43                                                               & 22.5                                                              & 1.42                                                                & 2.5                                                                \\ \cline{2-12} 
                                       & Coarse                               & \textbf{80.1}                                                            & \textbf{97.69}                                                                  & 9.05                                                               & 7.79                                                               & 87.94                                                              & 80.33                                                              & 3.02                                                               & 10.66                                                             & 0                                                                   & 1.23                                                               \\ \hline \hline
\multirow{2}{*}{\textbf{LARGE COEFF}}  & Fine                                 & 70.1                                                                     & 96.59                                                                           & 13.38                                                              & 10                                                                 & 69.9                                                               & 59.32                                                              & 13.38                                                              & 25.45                                                             & 3.34                                                                & 5.23                                                               \\ \cline{2-12} 
                                       & Coarse                               & \textbf{73.2}                                                            & \textbf{96.66}                                                                  & 10.45                                                              & 7.84                                                               & 79.85                                                              & 70.87                                                              & 7.84                                                               & 18.49                                                             & 1.87                                                                & 2.8                                                                \\ \hline \hline
\multirow{2}{*}{\textbf{NO BACKTRACK}} & Fine                                 & 46.5                                                                     & 92.93                                                                           & 9.16                                                               & 5.15                                                               & 74.21                                                              & 57.9                                                               & 14.58                                                              & 33.69                                                             & 2.06                                                                & 3.25                                                               \\ \cline{2-12} 
                                       & Coarse                               & \textbf{65.5}                                                            & \textbf{95.7}                                                                   & 3.19                                                               & 2.61                                                               & 90.14                                                              & 77.31                                                              & 5.22                                                               & 18.27                                                             & 1.45                                                                & 1.81                                                               \\ \hline \hline
\multirow{2}{*}{\textbf{MEDIUM DEG}}   & Fine                                 & 59.6                                                                     & 95.28                                                                           & 7.43                                                               & 5.48                                                               & 72.03                                                              & 57.1                                                               & 15.35                                                              & 32.9                                                              & 5.2                                                                 & 4.52                                                               \\ \cline{2-12} 
                                       & Coarse                               & \textbf{65.1}                                                            & \textbf{95.61}                                                                  & 6.88                                                               & 5.26                                                               & 78.51                                                              & 67.79                                                              & 11.46                                                              & 24.63                                                             & 3.15                                                                & 2.32                                                               \\ \hline \hline
\multirow{2}{*}{\textbf{MEDIUM TERMS}} & Fine                                 & \textbf{56.9}                                                            & \textbf{95.7}                                                                   & 21.58                                                              & 13.3                                                               & 67.29                                                              & 57.72                                                              & 8.58                                                               & 23.96                                                             & 2.55                                                                & 5.02                                                               \\ \cline{2-12} 
                                       & Coarse                               & 52.8                                                                     & 94.57                                                                           & 23.94                                                              & 15.6                                                               & 68.22                                                              & 62.77                                                              & 3.6                                                                & 16.67                                                             & 4.24                                                                & 4.96                                                               \\ \hline
\end{tabular}%
}
\caption{Errors for \textsc{fine} and  \textsc{coarse} configurations for 2 Variable for annotated proofs (No curriculum).}
\label{tab:annotate2varerror}
\end{table}

\section{Out-of-Distribution Evaluation}
\label{sec:appendx:ood}
We present the results for Out-of-Distribution evaluation here. Table \ref{tab:OOD2var} contains results for best 2 variable models (Prefix/Coarse) tested on 1 Variable setting.
\\ Table \ref{tab:OODcoeff} contains results for best 1 variable models (Prefix/Coarse) tested on \textsc{Small, medium} and \textsc{Large} coefficient setting. As expected, the \textsc{Small} and \textsc{medium} models perform much worse when tested on higher coefficients.
\\ We also evaluated the best 1 variable models (Prefix/Coarse) on \textsc{medium degree} and \textsc{terms} settings, to check generalization with respect to \# terms and degree of polynomial. Table \ref{tab:OODterm} contains results for the same. The \textsc{Medium Coeff} model is not able to generalize to more terms or polynomials of higher degree.

\begin{table}[!htpb]
\centering
\resizebox{0.8\textwidth}{!}{%
\begin{tabular}{|c|c|c|c|c|c|c|}
\hline
\multirow{2}{*}{\textbf{Config}}                                  & \multicolumn{2}{c|}{\textbf{\begin{tabular}[c]{@{}c@{}}Train/Test= 2 Var/1 Var\end{tabular}}}                                             & \multicolumn{2}{c|}{\textbf{\begin{tabular}[c]{@{}c@{}}Train/Test= 1 Var/1 Var\end{tabular}}}                                             & \multicolumn{2}{c|}{\textbf{\begin{tabular}[c]{@{}c@{}}Train/Test= 2 Var/2 Var\end{tabular}}}                                             \\ \cline{2-7} 
                                                                  & \textbf{\begin{tabular}[c]{@{}c@{}}Full \\ Proof Acc.\end{tabular}} & \textbf{\begin{tabular}[c]{@{}c@{}}Greedy\\  Stepwise Acc.\end{tabular}} & \textbf{\begin{tabular}[c]{@{}c@{}}Full \\ Proof Acc.\end{tabular}} & \textbf{\begin{tabular}[c]{@{}c@{}}Greedy\\  Stepwise Acc.\end{tabular}} & \textbf{\begin{tabular}[c]{@{}c@{}}Full \\ Proof Acc.\end{tabular}} & \textbf{\begin{tabular}[c]{@{}c@{}}Greedy\\  Stepwise Acc.\end{tabular}} \\ \hline
\textbf{\begin{tabular}[c]{@{}c@{}}SMALL \\ COEFF\end{tabular}}   & \textbf{95.34}                                                          & \textbf{99.12}                                                              & 95.3                                                                    & 98.97                                                                       & 91.2                                                                    & 98.08                                                                       \\ \hline
\textbf{\begin{tabular}[c]{@{}c@{}}MEDIUM \\ COEFF\end{tabular}}  & 87.4                                                                    & 97.11                                                                       & \textbf{93.6}                                                           & \textbf{98.58}                                                              & 84.5                                                                    & 96.03                                                                       \\ \hline
\textbf{\begin{tabular}[c]{@{}c@{}}LARGE \\ COEFF\end{tabular}}   & \textbf{89.4}                                                           & \textbf{97.13}                                                              & 83.5                                                                    & 96.25                                                                       & 83.7                                                                    & 96.23                                                                       \\ \hline
\textbf{\begin{tabular}[c]{@{}c@{}}NO  BACK\\ TRACK\end{tabular}} & \textbf{84.2}                                                           & \textbf{98.29}                                                              & 79.7                                                                    & 95.38                                                                       & 63.2                                                                    & 89.87                                                                       \\ \hline
\textbf{\begin{tabular}[c]{@{}c@{}}MEDIUM \\ DEG\end{tabular}}    & \textbf{87.7}                                                           & \textbf{97.83}                                                              & 87.7                                                                    & 96.82                                                                       & 83.4                                                                    & 96.41                                                                       \\ \hline
\textbf{\begin{tabular}[c]{@{}c@{}}MEDIUM \\ TERMS\end{tabular}}  & \textbf{78.5}                                                           & \textbf{96.16}                                                              & 76.3                                                                    & 95.78                                                                       & 67.8                                                                    & 93.58                                                                       \\ \hline
\end{tabular}%
}
\caption{Results for OOD Testing. NVAR = 2 \textsc{Coarse/Prefix} models tested on corresponding NVAR = 1 setting (No curriculum, No annotation).}
\label{tab:OOD2var}
\end{table}

\begin{table}[!htpb]
\centering
\resizebox{0.8\textwidth}{!}{%
\begin{tabular}{|c|c|c|c|c|c|c|}
\hline
\multirow{3}{*}{\textbf{\begin{tabular}[c]{@{}c@{}}Train\\ Config\end{tabular}}} & \multicolumn{6}{c|}{\textbf{Test Config}}                                                                                                                                                                                                                                                                                                                                                                       \\ \cline{2-7} 
                                                                                 & \multicolumn{2}{c|}{\textbf{\begin{tabular}[c]{@{}c@{}}SMALL \\ COEFF\end{tabular}}}                                                & \multicolumn{2}{c|}{\textbf{\begin{tabular}[c]{@{}c@{}}MEDIUM \\ COEFF\end{tabular}}}                                               & \multicolumn{2}{c|}{\textbf{\begin{tabular}[c]{@{}c@{}}LARGE \\ COEFF\end{tabular}}}                                                \\ \cline{2-7} 
                                                                                 & \begin{tabular}[c]{@{}c@{}}Full \\ Proof \\  Acc.\end{tabular} & \begin{tabular}[c]{@{}c@{}}Greedy\\  Stepwise\\  Acc.\end{tabular} & \begin{tabular}[c]{@{}c@{}}Full \\ Proof \\  Acc.\end{tabular} & \begin{tabular}[c]{@{}c@{}}Greedy\\  Stepwise\\  Acc.\end{tabular} & \begin{tabular}[c]{@{}c@{}}Full \\ Proof \\  Acc.\end{tabular} & \begin{tabular}[c]{@{}c@{}}Greedy\\  Stepwise\\  Acc.\end{tabular} \\ \hline
\textbf{\begin{tabular}[c]{@{}c@{}}SMALL \\ COEFF\end{tabular}}                  & 95.3                                                           & 98.97                                                              & 33.4                                                           & 69.05                                                              & 31                                                             & 68.02                                                              \\ \hline
\textbf{\begin{tabular}[c]{@{}c@{}}MEDIUM \\ COEFF\end{tabular}}                 & \textbf{96.6}                                                  & \textbf{99.29}                                                     & 93.6                                                           & 98.58                                                              & 33.6                                                           & 68.96                                                              \\ \hline
\textbf{\begin{tabular}[c]{@{}c@{}}LARGE \\ COEFF\end{tabular}}                  & 95.8                                                           & 99.1                                                               & \textbf{94.4}                                                  & \textbf{98.64}                                                     & \textbf{83.5}                                                  & \textbf{96.25}                                                     \\ \hline
\end{tabular}%
}
\caption{OOD Testing: Prefix/Coarse 1 Variable Models tested on various coefficient limit configurations (\textsc{Small, Medium} and \textsc{Coarse}). (No curriculum, No annotation).}
\label{tab:OODcoeff}
\end{table}

\begin{table}[!htpb]
\centering
\resizebox{0.8\textwidth}{!}{%
\begin{tabular}{|c|c|c|c|c|c|c|}
\hline
\multirow{3}{*}{\textbf{\begin{tabular}[c]{@{}c@{}}Train\\ Config\end{tabular}}} & \multicolumn{6}{c|}{\textbf{Test Config}}                                                                                                                                                                                                                                                                                                                                                                       \\ \cline{2-7} 
                                                                                 & \multicolumn{2}{c|}{\textbf{\begin{tabular}[c]{@{}c@{}}MEDIUM \\ COEFF\end{tabular}}}                                               & \multicolumn{2}{c|}{\textbf{\begin{tabular}[c]{@{}c@{}}MEDIUM \\ DEG\end{tabular}}}                                                 & \multicolumn{2}{c|}{\textbf{\begin{tabular}[c]{@{}c@{}}MEDIUM \\ TERMS\end{tabular}}}                                               \\ \cline{2-7} 
                                                                                 & \begin{tabular}[c]{@{}c@{}}Full \\ Proof \\  Acc.\end{tabular} & \begin{tabular}[c]{@{}c@{}}Greedy\\  Stepwise\\  Acc.\end{tabular} & \begin{tabular}[c]{@{}c@{}}Full \\ Proof \\  Acc.\end{tabular} & \begin{tabular}[c]{@{}c@{}}Greedy\\  Stepwise\\  Acc.\end{tabular} & \begin{tabular}[c]{@{}c@{}}Full \\ Proof \\  Acc.\end{tabular} & \begin{tabular}[c]{@{}c@{}}Greedy\\  Stepwise\\  Acc.\end{tabular} \\ \hline
\textbf{\begin{tabular}[c]{@{}c@{}}MEDIUM \\ COEFF\end{tabular}}                 & 93.6                                                           & 98.58                                                              & 20.8                                                           & 47.77                                                              & 26.1                                                           & 54.65                                                              \\ \hline
\textbf{\begin{tabular}[c]{@{}c@{}}MEDIUM \\ DEG\end{tabular}}                   & 94.8                                                           & 98.93                                                              & 87.7                                                           & 96.82                                                              & 25.5                                                           & 54.39                                                              \\ \hline
\textbf{\begin{tabular}[c]{@{}c@{}}MEDIUM \\ TERMS\end{tabular}}                 & 92.7                                                           & 96.87                                                              & 18.6                                                           & 46.97                                                              & 76.3                                                           & 95.78                                                              \\ \hline
\end{tabular}%
}
\caption{OOD Testing: Prefix/Coarse 1 Variable Models tested on various \#term and degree configurations (\textsc{Medium degree} and \textsc{Medium terms}). (No curriculum, No annotation).}
\label{tab:OODterm}
\end{table}

\section{Curriculum Learning}
\label{sec:appendx:curr}
Learning the simplification steps should entail learning the sub-tasks, such as addition and  multiplication (of numeric coefficients and symbolic variables); where multiplying variables precludes learning to add exponents of similar variables. As these sub-tasks are well-defined and dependencies among them are clear, we explore different types of curriculums based on the Mastering-Rate-based (MR) curriculum learning algorithm proposed in \cite{willems2020mastering}. Authors in \cite{willems2020mastering} define curriculum learning by 1) a \textit{curriculum} i.e. a set of tasks $\mathcal{C}=\{c_1, \ldots, c_n\}$, where a task is set of examples of similar type with a sampling distribution, and 2) a \textit{program} which for each training step defines the tasks to train the learner given its learning state and the curriculum. Formally, the program $d: \mathbb{N} \rightarrow \mathcal{D}^{\mathcal{C}}$, is a sequence of distributions over $\mathcal{C}$.
The authors estimate the \textit{program} function through an \textit{attention} function which defines attention over the tasks at a time-step, and an \textit{attention-to-distribution converter} which converts the attention to a distribution over $\mathcal{C}$. Authors observe that other algorithms \citep{matiisen2019teacher,graves2017automated} are special cases of the above setting with different choices for \textit{program}. 

To learn on tasks that are \textit{learnable but not learnt yet}, authors define an \textit{ordered curriculum} $\mathcal{O}^\mathcal{C}$ which is a directed graph over tasks in $\mathcal{C}$. An edge from A to B indicates that learning task A before B is preferable. For supervised learners, the learnability for each task depends on mastering rate ($\mathcal{M}_c(t)$) computed from the normalized mean accuracy for that task at time-step $t$. At each time-step, the MR algorithm computes attention over a task ($a_c(t)$) from mastering rates of its ancestors and successors. During training to sample batches, a hyperparameter $N_b$ for the curriculum determines the number of batches to be considered at a step, before re-computing the  attention over tasks. Using the \textit{program} $d$, we first sample $N_b*b$ examples from tasks in $\mathcal{C}$. The model is then trained on randomly sampled $N_b$ minibatches are sampled updating the mastering rates.

For polynomial simplification for 1 variable, we define the following tasks \textsc{Add}, \textsc{Mul2}, \textsc{Mul3}, \textsc{Scoeff} and \textsc{Mixed}. For \textsc{Add}, only one factor per product is allowed, so there is no multiplication. For \textsc{Mul2} and \textsc{Mul3} only 1 product is allowed with maximum two factors and three factors respectively. \textsc{Scoeff} points to the \textsc{Small Coeff} configuration and \textsc{Mixed} is the final variable size configuration of the target variable configuration. We define the following curriculums:
\begin{itemize}
    \item  C: \{{\sc (Add, Mul3), (Mul3, Mixed), (Add, Mixed)}\}.
    \item C2: \{{\sc (Add, Mul2), (Mul2, Mul3), (Mul3, Mixed), (Add, Mixed)}\}.
    \item C4: \{{\sc (Add, Mul2), (Mul2, Mul3), (Mul3, Scoeff), (Add, Scoeff) (Scoeff, Mixed)}\}.
\end{itemize}
For all our experiments, we use the MR algorithm with gAmax Linreg A2D converter functions described in \cite{willems2020mastering}. Model parameters and the training configurations remains the same as before\footnote{We use $N_b$ as 10. For other default parameters in CL, please check \url{github.com/lcswillems/automatic-curriculum}.}. We show the results in Table \ref{tab:cl:var1coarse} for \textsc{Coarse} configuration.   As coefficient size grows from \textsc{Small}, \textsc{Medium}, \textsc{Large} to \textsc{No Backtrack} - the improvements in full proof accuracy steadily increase from $1\%$ to $10.84\%$. For \textsc{No Backtrack}, the improvement in top-1 accuracy is by $20\%$ from a no curriculum setting.  However, we observe for \textsc{Medium Terms}, there is a drop in accuracy for all curriculums and input representations. It is possible that, more carefully designed curriculums may improve the results. There is no conceivable pattern observed for infix or prefix representations. However, compared to learning without curriculum, the improvement observed for infix representation is larger than prefix.

\begin{table}[!htpb]
\centering
\resizebox{\textwidth}{!}{%
\begin{tabular}{|l|l|l|c|c|c|c|c|c|c|c|c|}
\hline
                                                                          &                          &                                                       & \multicolumn{1}{l|}{}        & \multicolumn{2}{c|}{Full Proof (Beam-1)}                                                                                                                                       & \multicolumn{2}{c|}{Step-wise (Beam-5)}                                                                                                                       & \multicolumn{4}{c|}{Calibration (Beam-5)}                                                                                                                                     \\ \hline
                                                                          &                          & \begin{tabular}[c]{@{}l@{}}Curri\\ culum\end{tabular} & \multicolumn{1}{l|}{\#Train} & \multicolumn{1}{l|}{\begin{tabular}[c]{@{}l@{}}Full Proof\\ Accuracy\end{tabular}} & \multicolumn{1}{l|}{\begin{tabular}[c]{@{}l@{}}Stepwise\\ Accuracy\end{tabular}} & \multicolumn{1}{l|}{\begin{tabular}[c]{@{}l@{}}Top-1\\ Acc\end{tabular}} & \multicolumn{1}{l|}{\begin{tabular}[c]{@{}l@{}}Beam-5\\ Acc\end{tabular}} & \multicolumn{1}{l|}{\begin{tabular}[c]{@{}l@{}}Sure\\ Rate\end{tabular}} & \multicolumn{1}{l|}{P}       & \multicolumn{1}{l|}{R}       & \multicolumn{1}{l|}{F1}     \\ \hline
                                                                          &                          & C                                                     & {2.8M}  & {94.38}                                                       & {98.76}                                                     & {94.84}                                             & {96.68}                                              & {89.36}                                             & {100}   & {94.22} & {0.97} \\ \cline{3-12} 
                                                                          & \multirow{-2}{*}{Infix}  & C2                                                    & {2M}    & {\textbf{95.98}}                                              & {\textbf{99.0}}                                             & {91.64}                                             & {93.24}                                              & {86.16}                                             & {99.9}  & {93.98} & { 0.97} \\ \cline{2-12} 
                                                                          &                          & C                                                     & {2.02M} & { 94.26}                                                       & { 98.65}                                                     & {77.76}                                             & {80.46}                                              & {70.62}                                             & {99.94} & { 90.77} & { 0.95} \\ \cline{3-12} 
\multirow{-4}{*}{\begin{tabular}[c]{@{}l@{}}Small \\ Coeff\end{tabular}}  & \multirow{-2}{*}{Prefix} & C2                                                    & { 2.29M} & { 94.6}                                                        & { 98.56}                                                     & { 93.44}                                             & { 95.28}                                              & { 88.02}                                             & { 99.89} & { 94.09} & { 0.97} \\ \hline
                                                                          &                          & C2                                                    & { 3.9M}  & { \textbf{95.44}}                                              & { \textbf{99.02}}                                            & { 94.86}                                             & { 96.44}                                              & { 91.18}                                             & { 100}   & { 96.12} & { 0.98} \\ \cline{3-12} 
                                                                          & \multirow{-2}{*}{Infix}  & C4                                                    & { 2M}    & { 93.86}                                                       & { 98.59}                                                     & { 88.22}                                             & { 90.24}                                              & { 84.68}                                             & { 99.91} & { 95.90} & { 0.98} \\ \cline{2-12} 
                                                                          &                          & C2                                                    & { 3.7M}  & { 94.78}                                                       & { 98.82}                                                     & { 91.98}                                             & { 93.66}                                              & { 88.08}                                             & { 99.93} & { 95.69} & { 0.98} \\ \cline{3-12} 
\multirow{-4}{*}{\begin{tabular}[c]{@{}l@{}}Medium \\ Coeff\end{tabular}} & \multirow{-2}{*}{Prefix} & C4                                                    & { 4.4M}  & { 94.8}                                                        & { 98.87}                                                     & { 85.3}                                              & { 87.82}                                              & { 80.62}                                             & { 99.98} & { 94.49} & { 0.97} \\ \hline
                                                                          &                          & C2                                                    & { 6.9M}  & { 91.26}                                                       & { 97.92}                                                     & { 96.4}                                              & { 98.06}                                              & { 90.24}                                             & { 99.89} & { 93.51} & { 0.97} \\ \cline{3-12} 
                                                                          & \multirow{-2}{*}{Infix}  & C4                                                    & { 7.6M}  & { 91.62}                                                       & { 98.16}                                                     & { 91.54}                                             & { 93.3}                                               & { 87.38}                                             & { 99.84} & { 95.3}  & { 0.98} \\ \cline{2-12} 
                                                                          &                          & C2                                                    & { 6.5M}  & { 92.2}                                                        & { 98.31}                                                     & { 85.38}                                             & { 87.78}                                              & { 81.42}                                             & { 99.95} & { 95.32} & { 0.98} \\ \cline{3-12} 
\multirow{-4}{*}{\begin{tabular}[c]{@{}l@{}}Large \\ Coeff\end{tabular}}  & \multirow{-2}{*}{Prefix} & C4                                                    & { 6.97M} & { \textbf{92.46}}                                              & { \textbf{98.42}}                                            & { 91.3}                                              & { 93.34}                                              & { 87.54}                                             & { 100.0} & { 95.88} & { 0.98} \\ \hline
                                                                          &                          & C2                                                    & { 4.8M}  & { 86.44}                                                       & { 97.27}                                                     & { 93.68}                                             & { 95.46}                                              & { 88.72}                                             & { 99.98} & { 94.68} & { 0.97} \\ \cline{3-12} 
                                                                          & \multirow{-2}{*}{Infix}  & C4                                                    & { 5.1M}  & { 85.96}                                                       & { 97.21}                                                     & { 94.64}                                             & { 96.1}                                               & { 89.5}                                              & { 100}   & { 94.57} & { 0.97} \\ \cline{2-12} 
                                                                          &                          & C2                                                    & { 7M}    & { 86.16}                                                       & { 97.30}                                                     & { 82.24}                                             & { 84.44}                                              & { 77.46}                                             & { 99.95} & { 94.14} & { 0.97} \\ \cline{3-12} 
\multirow{-4}{*}{\begin{tabular}[c]{@{}l@{}}No\\ Backtrack\end{tabular}}  & \multirow{-2}{*}{Prefix} & C4                                                    & { 5.5M}  & { \textbf{86.48}}                                              & { \textbf{97.45}}                                            & { 92.6}                                              & { 94.3}                                               & { 87.78}                                             & { 99.95} & { 94.75} & { 0.97} \\ \hline
                                                                          &                          & C2                                                    & { 3.5M}      & { 87.12}                                                   & { 97.01}                                                 & { 84.16}                                                  & { 87.44}                                                   & { 78.46}                                                  & { 99.95}      & { 93.18}      & { 0.96}     \\ \cline{3-12} 
                                                                          & \multirow{-2}{*}{Infix}  & C4                                                    & { 3.4M}  & { 94.12}                                                       & { 98.65}                                                     & { 90.62}                                             & { 81.984}                                             & { 86.66}                                             & { 99.93} & { 95.56} & { 0.98} \\ \cline{2-12} 
                                                                          &                          & C2                                                    & { 5.35M} & { \textbf{94.28}}                                              & { \textbf{98.71}}                                            & { 80.8}                                              & { 82.84}                                              & { 75.76}                                             & { 100}   & { 93.51} & { 0.97} \\ \cline{3-12} 
\multirow{-4}{*}{\begin{tabular}[c]{@{}l@{}}Medium\\ Degree\end{tabular}} & \multirow{-2}{*}{Prefix} & C4                                                    & { 3.5M}  & { 92.38}                                                       & { 98.30}                                                     & { 83.7}                                              & { 85.48}                                              & { 78.94}                                             & { 99.92} & { 94.24} & { 0.97} \\ \hline
                                                                          &                          & C2                                                    & { 4.4M}  & { 59.54}                                                       & { 75.76}                                                     & { 65.6}                                              & { 69.56}                                              & { 60.84}                                             & { 95.36} & { 88.45} & { 0.92} \\ \cline{3-12} 
                                                                          & \multirow{-2}{*}{Infix}  & C4                                                    & { 3.8M}  & { 56.94}                                                       & { 76.72}                                                     & { 69.84}                                             & { 73.44}                                              & { 60.76}                                             & { 97.5}  & { 84.82} & { 0.91} \\ \cline{2-12} 
                                                                          &                          & C2                                                    & { 2.8M}  & { 41.84}                                                       & { 51.24}                                                     & { 40.62}                                             & { 45.36}                                              & { 36.9}                                              & { 92.57} & { 84.10} & { 0.88} \\ \cline{3-12} 
\multirow{-4}{*}{\begin{tabular}[c]{@{}l@{}}Medium\\ Terms\end{tabular}}  & \multirow{-2}{*}{Prefix} & C4                                                    & { 3.37M} & { 49.02}                                                       & { 65.41}                                                     & { 58.56}                                             & { 64.64}                                              & { 45.44}                                             & { 96.83} & { 75.14} & { 0.85} \\ \hline
\end{tabular}%
}
\caption{Curriculum Learning results for 1 variable for the \textsc{Coarse} configuration for both Infix and prefix representations.}
\label{tab:cl:var1coarse}
\end{table}




\end{document}